\documentclass[10pt,twocolumn,letterpaper]{article}

\usepackage[pagenumbers]{cvpr} % To force page numbers, e.g. for an arXiv version
\definecolor{cvprblue}{rgb}{0.21,0.49,0.74}
\usepackage[pagebackref,breaklinks,colorlinks,allcolors=cvprblue]{hyperref}
\usepackage{comment}
\usepackage{mdframed}
\usepackage{multirow}
\usepackage{caption}
\usepackage{cuted}
\usepackage[table]{xcolor}
\usepackage{tabularx}
\usepackage{xcolor}
\newlength\savewidth
\newcommand\shline{\noalign{\global\savewidth\arrayrulewidth
		\global\arrayrulewidth 1.1pt}%
	\hline
	\noalign{\global\arrayrulewidth\savewidth}}
\def\paperID{5806} % *** Enter the Paper ID here
\def\confName{CVPR}
\def\confYear{2026}

\title{Learning to Reason in 4D: Dynamic Spatial Understanding \\for Vision Language Models\vspace{-15pt}}
\author{Shengchao Zhou\textsuperscript{\rm 1}, Yuxin Chen\textsuperscript{\rm 2}, Yuying Ge\textsuperscript{\rm 2}, Wei Huang\textsuperscript{\rm 1}, Jiehong Lin\textsuperscript{\rm 1}, Ying Shan\textsuperscript{\rm 2} and Xiaojuan Qi\textsuperscript{\rm 1}\setcounter{footnote}{1}\thanks{Corresponding author.} 
\\
\textsuperscript{\rm 1}The University of Hong Kong~~~~~~~~~~~~\textsuperscript{\rm 2}ARC Lab, Tencent PCG\\
{\tt\small \{zhoushengchao2024, weih\}@connect.hku.hk, \{uasonchen, yingsshan\}@tencent.com}\\
{\tt\small \{yyge13, mortimer.jh.lin\}@gmail.com, xjqi@eee.hku.hk
}\\
\textcolor[HTML]{347CBC}{https://github.com/TencentARC/DSR\_Suite}
}

\begin{document}

\maketitle
\setlength{\stripsep}{-0.75cm}
\begin{strip}
\centering
\setlength{\belowcaptionskip}{1.1cm}
\includegraphics[width=0.972\textwidth]{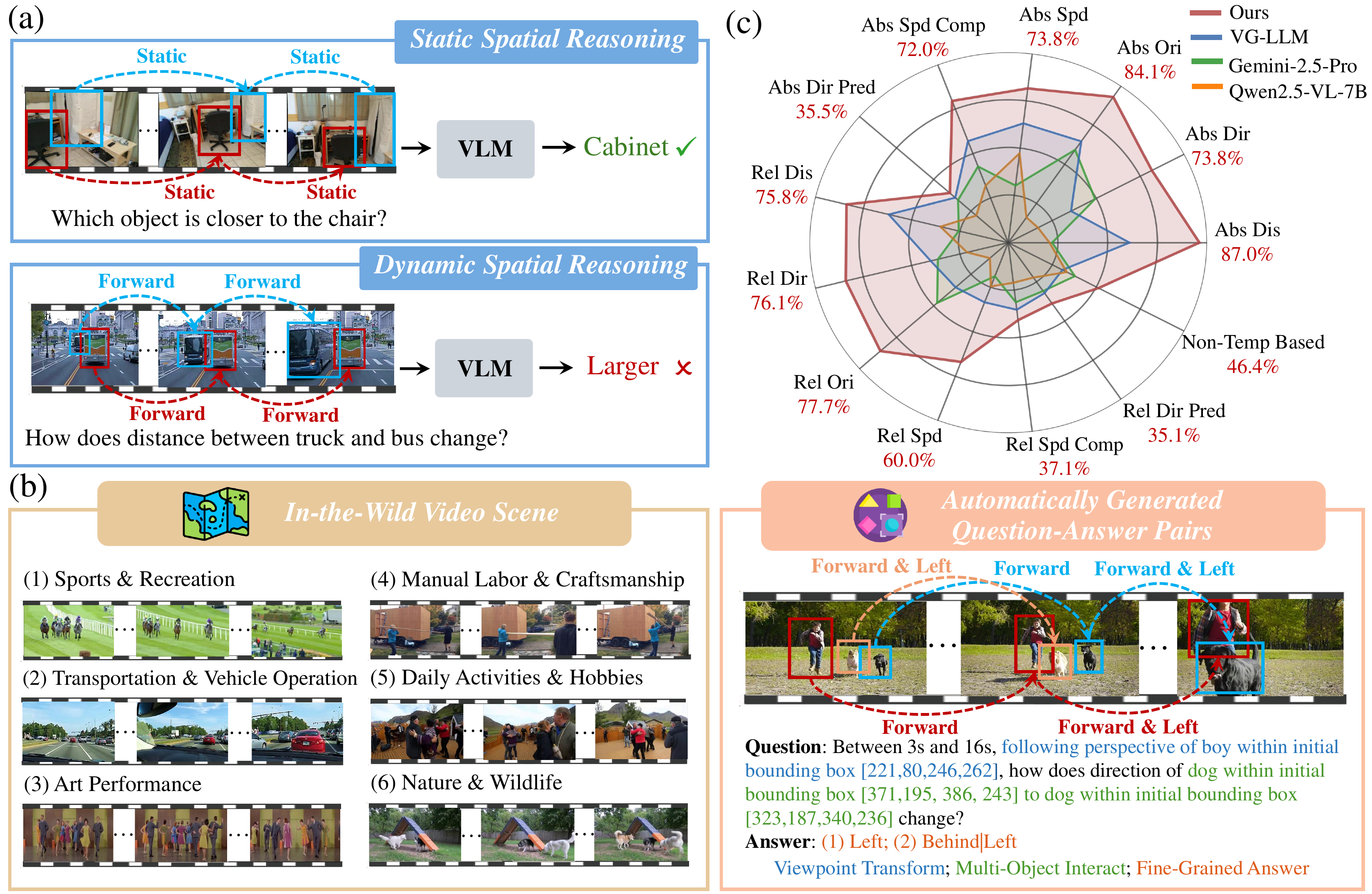}
\captionof{figure}{\textbf{Overview of DSR Suite}. (a) Comparison between static and dynamic spatial reasoning: Unlike static scenarios, dynamic spatial reasoning (DSR) requires understanding environments with moving objects, posing greater reasoning challenges. (b) Data examples from DSR Suite: Built upon automated pipeline and in-the-wild videos, DSR Suite generates scalable question-answer pairs that feature viewpoint transform, multi-object interact and fine-grained answers for comprehensive training and evaluation of DSR. (c) Benchmark comparison: Evaluations on our proposed DSR-Bench highlight the capability of our model trained on constructed DSR-Train.}
\label{img:teaser}
\end{strip}

\begin{abstract}

Vision-language models (VLM) excel at general understanding yet remain weak at dynamic spatial reasoning (\textbf{DSR}), i.e., reasoning about the evolvement of object geometry and relationship in 3D space over time, largely due to the scarcity of scalable 4D-aware training resources. To bridge this gap across aspects of dataset, benchmark and model, we introduce \textbf{DSR Suite}. First, we propose an automated pipeline that generates multiple-choice question-answer pairs from in-the-wild videos for DSR. By leveraging modern vision foundation models, the pipeline extracts rich geometric and motion information, including camera poses, local point clouds, object masks, orientations, and 3D trajectories. These geometric cues enable the construction of \textbf{DSR-Train} for learning and further human-refined \textbf{DSR-Bench} for evaluation. Compared with previous works, our data emphasize (i) in-the-wild video sources, (ii) object- and scene-level 3D requirements, (iii) viewpoint transformations, (iv) multi-object interactions, and (v) fine-grained, procedural answers. Beyond data, we propose a lightweight Geometry Selection Module (\textbf{GSM}) to seamlessly integrate geometric priors into VLMs, which condenses question semantics and extracts question-relevant knowledge from pretrained 4D reconstruction priors into a compact set of geometry tokens. This targeted extraction avoids overwhelming the model with irrelevant knowledge. Experiments show that integrating DSR-Train and GSM into Qwen2.5-VL-7B significantly enhances its dynamic spatial reasoning capability, while maintaining accuracy on general video understanding benchmarks.

\end{abstract}    
\section{Introduction}
\label{sec:intro}

Recent advances in vision-language models (VLMs) have led to remarkable progress in video understanding, driven by large-scale multimodal pretraining and alignment between visual and textual representations \cite{lillava,chenlongvila,zhang2023video,lin2024video,weng2024longvlm}. Despite these achievements, VLMs remain limited to perform 3D spatial reasoning in dynamic environments \cite{wangcompositional}, which requires understanding how the object geometry and relationship evolve in 3D space over time, \ie, dynamic spatial reasoning (DSR). This capability is fundamental for interactive AI systems in robotics, autonomy, AR/VR and embodied intelligence, where the environments are dynamic and spatial relationships constantly evolve.

Recent works have started to explore 3D spatial reasoning within VLMs \cite{chen2024spatialvlm,cheng2024spatialrgpt,zhu2024llava}. However, as in Figure \ref{img:teaser}(a), most efforts remain constrained to static scenes with immobile objects \cite{yang2025thinking,zhang2025flatland,cai2025spatialbot} or short-horizon motion \cite{yang2025mmsi,wu2025spatialscore,ray2024sat,han2025videoespresso}, providing limited insight into 4D intelligence. Only a handful of studies consider dynamic scenes \cite{zhou2025vlm4d,li2025sti}, yet they suffer from narrow scene diversity, limited reasoning types and lack of training data. Beyond data, current methods typically inject geometric knowledge to VLMs through direct cross-attention or naïve fusion with a large set of pretrained reconstructive priors \cite{fan2025vlm,zheng2025learning,zheng2025video}, compromising general-purpose performance by overfitting to task-specific cues. Consequently, a scalable pipeline, training corpus, comprehensive benchmark and effective model for systematically investigating DSR is in demand.

To this end, we present \textbf{DSR Suite}, a framework designed to endow VLMs with DSR ability. At its core, DSR Suite introduces an automated data-generation pipeline that constructs training and evaluation datasets. The pipeline curates in-the-wild videos and leverages modern vision foundation models to extract camera poses, local point clouds, object masks, orientations and 3D trajectories. As in Figure \ref{img:teaser}(b), with these annotations, we derive: (1) \textbf{DSR-Train}, a large-scale multiple-choice corpus for DSR training; and (2) \textbf{DSR-Bench}, a further human-refined benchmark that evaluates object- and scene-level 3D understanding, multi-object interactions, viewpoint transformations, and fine-grained temporal reasoning in dynamic scenes. Together, these resources bridge a critical gap for studying geometric and temporal reasoning in realistic, dynamic settings.

Beyond data, we further propose a simple yet effective mechanism to integrate geometric priors from pretrained 3D foundation models into VLMs. While training on DSR-Train through naïvely adding a large set of features from 3D reconstruction models to vision tokens can improve DSR performance, it will lead to noticeable degradation on general benchmarks (see Table \ref{tab:2} in Section \ref{sec:abl_1}). To mitigate this trade-off, we propose a lightweight Geometry Selection Module (GSM) that performs text-guided geometric knowledge selection. Concretely, GSM adopts a dual Q-Former design: the first condenses question semantics, and the second extracts only question-relevant knowledge from 4D reconstruction priors into a compact set of geometry tokens. This targeted filtering suppresses irrelevant geometric noise, enabling precise 4D grounding without compromising general multimodal understanding performance.

Our contributions can be summarized as follows: 
\begin{itemize}
 \item We build an automated pipeline that transforms in-the-wild videos into multiple-choice question-answer pairs for dynamic spatial reasoning by extracting camera poses, local point clouds, object masks, orientations and 3D trajectories, yielding DSR-Train for model training and DSR-Bench for comprehensive evaluation. 
 %We introduce an automated multi-choice video question answering data generation pipeline targeting at 3D spatial reasoning in dynamic scenes. Based on this pipeline, we construct both training dataset DSR-Train and more comprehensive evaluation benchmark DSR-Bench.
\item We propose GSM, a lightweight module that integrates geometric priors into VLMs via two stacked Q-Formers: the first condenses question semantics, and the second extracts question-relevant knowledge from 4D reconstruction priors into a compact set of geometry tokens, mitigating effect of irrelevant noise for general understanding. 
%To mitigate influence of irrelevant 3D knowledge, we propose GATE, which is a module that integrates 3D knowledge to VLMs in a different way. With two stacked Q-Formers, GATE sequentially condenses question and extracts question-relevant 3D knowledge into a small set of queries to be appended with vision tokens.
\item Experiments show that after training on DSR-Train, Qwen2.5-VL-7B with GSM achieves state-of-the-art performance on DSR-Bench, while maintaining competitive results across general multimodal benchmarks. 
% Yx:
% achieves state-of-the-art performance on DSR-Bench, while maintaining competitive results across general multimodal benchmarks.
\end{itemize}
\section{Related Work}
\label{sec:rel_work}
%-------------------------------------------------------------------------
\textbf{Spatial Reasoning Data} Spatial reasoning aims to endow VLMs with the ability to reason in 3D space, thereby extending their applicability to more downstream tasks such as robotic navigation. To foster this capability, a growing number of training datasets and benchmarks have recently been proposed. Among those using multi-image or video input, training data in \cite{zhang2025flatland} and benchmark in \cite{yang2025thinking} present early attempts that demand aggregation of spatial information across frames. Despite these advances, most existing works are restricted to static scenes, where objects are motionless. In contrast, real-world environments are inherently dynamic, with objects moving and changing positions.

Several studies attempt to incorporate object motion into their benchmark designs. In \cite{yang2025mmsi,wu2025spatialscore,ray2024sat}, VLMs are provided with two images to infer object changes. However, this short temporal horizon limits the evaluation of long-term reasoning ability. Subsequent efforts \cite{zhou2025vlm4d,li2025sti,jia2025omnispatial} extend input modality to videos, yet these datasets are confined primarily to autonomous driving or human–object interaction scenarios. Moreover, they often suffer from limited question diversity, weak 3D understanding requirements, and coarse-grained answer formulations, hindering comprehensive evaluation of DSR. Beyond these benchmarks, there is still a lack of datasets for task training, leaving current models without sufficient supervision.

In contrast, leveraging our automated data generation pipeline, we construct both a training dataset, DSR-Train, and an evaluation benchmark, DSR-Bench, that overcome these limitations. Based on in-the-wild videos, diverse question types, randomized target object and viewpoint, and fine-grained answers, these resources enables comprehensive and fine-grained training and evaluation of dynamic spatial reasoning in realistic settings. A detailed comparison with existing benchmarks is presented in Section \ref{sec:3.4}.

\vspace{0.05in}\noindent\textbf{Spatial Reasoning Models}
To enhance spatial reasoning capabilities, several methods are proposed to integrate 3D information into VLMs. In \cite{zhu2024llava}, the authors encode 3D coordinates of each pixel as position embeddings to provide geometric context for downstream tasks. In \cite{fan2025vlm,zheng2025learning}, knowledge extracted from 3D foundation models, such as CUT3R \cite{wang2025continuous} and VGGT \cite{wang2025vggt}, are directly fused with vision tokens. While these approaches introduce explicit geometric priors, the injected features are often task-specific and may lead to performance degradation on benchmarks unrelated to spatial reasoning. In contrast, our GSM selectively extracts question-relevant 3D knowledge into a compact set of tokens, thereby reducing interference with general multimodal understanding while strengthening DSR ability.
\begin{figure*}[t]
\centering
\includegraphics[width=\linewidth]{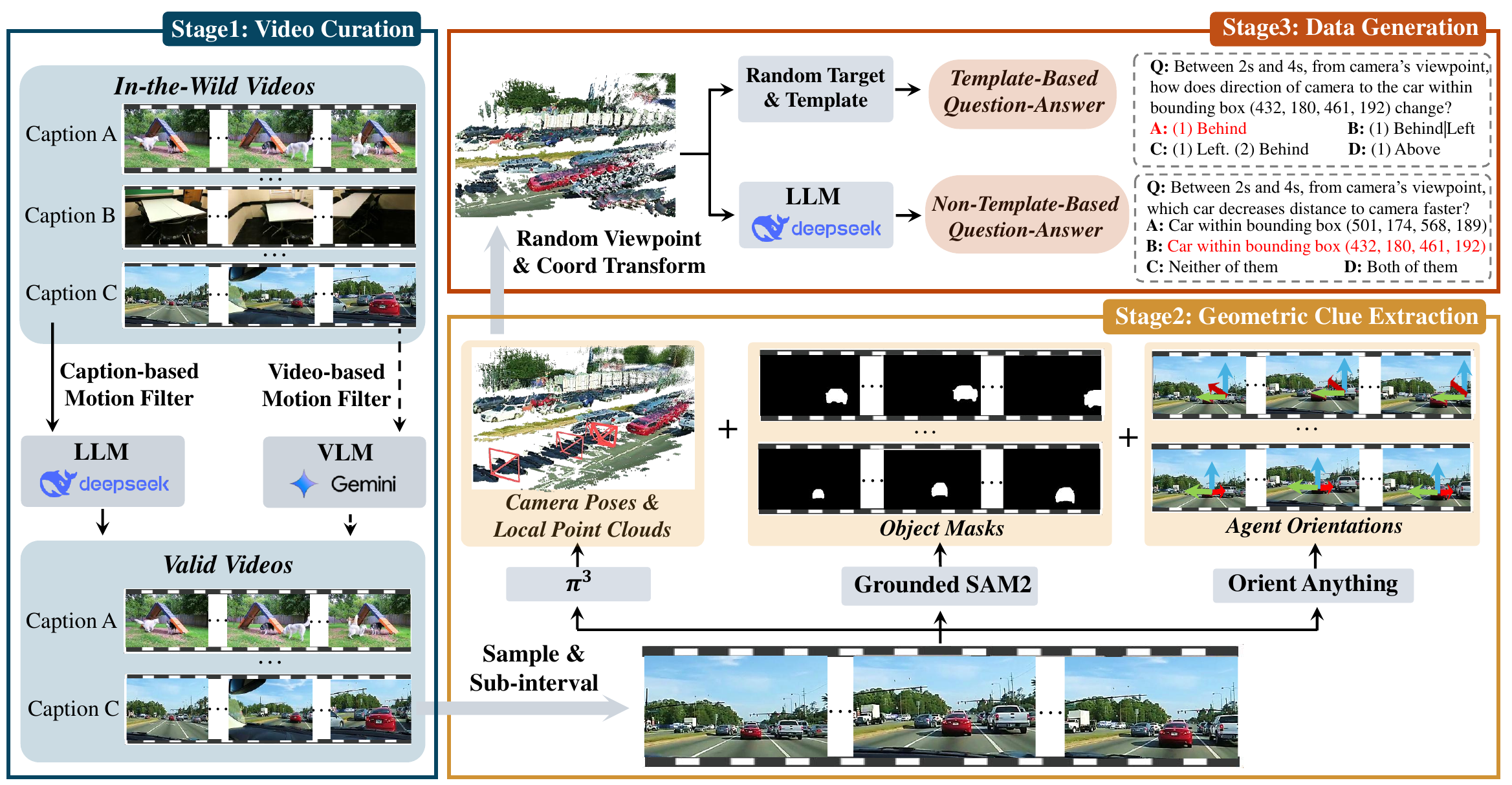}
\caption{Multiple-choice question–answer generation pipeline in our DSR Suite. It comprises three stages: \textbf{Video Curation}, \textbf{Geometric Clue Extraction} and \textbf{Data Generation}. In Video Curation stage, in-the-wild videos are filtered by LLMs or VLMs to remove motionless ones based on captions or visual cues. During Geometric Clue Extraction, vision foundation models extract key geometric cues, including camera poses, point clouds, object masks and orientations. Finally, in Data Generation, object coordinates are transformed into a randomly selected viewpoint and question–answer pairs are produced using either predefined templates or LLM-based free-form generation.}
\label{img:2}
\end{figure*} 

\section{Dynamic Spatial Reasoning Data}

To study DSR at scale, our DSR Suite first builds a fully automated pipeline that constructs both a large-scale training set (\textbf{DSR-Train}) and a further human-refined evaluation benchmark (\textbf{DSR-Bench}) from in-the-wild videos. The pipeline consists of three stages as illustrated in Figure \ref{img:2}:

%Prof. Qi's Version
%\xjqi{we need a figure for the data processing pipeline, show the diversity of the datsets}

%To study 4D spatial reasoning at scale, we build a fully automated pipeline that constructs both a large-scale training set (\textbf{DSR-Train}) and a human-refined evaluation benchmark (\textbf{DSR-Bench}) from in-the-wild videos. The pipeline consists of three stages (\xjqi{refer to our figure ...}):
(1) \textbf{Video Curation}--selecting videos with meaningful object motion;
(2) \textbf{Geometric Clue Extraction}--extracting both object- and scene-level 3D cues with vision foundation models such as Grounded SAM2 \cite{ren2024grounded}, Orient Anything \cite{wangorient}, $\pi^3$ \cite{wang2025pi};
(3) \textbf{Data Generation}--constructing diverse multiple-choice questions and fine-grained answers. 
Note that for \textbf{DSR-Bench}, all question-answer pairs are further refined by human annotators to ensure accuracy.
Together, this pipeline yields scalable, high-quality data for training and evaluation of dynamic spatial reasoning.

%Here, we introduce DSR-Train, a large-scale dataset for the training of dynamic spatial reasoning, and DSR-Bench, a benchmark developed to comprehensively evaluate this capability. Both are constructed with a scalable, fully automated pipeline comprising three stages: (1) \textbf{Video Curation} that collects videos from in-the-wild dataset and filters out samples with negligible object motion; (2) \textbf{Geometric Clue Extraction} that applies vision foundation models to videos to extract camera poses, local point clouds, object masks and orientations; (3) \textbf{Data Generation} that constructs both template-based and non-template-based question-answer pairs with randomized target objects and viewpoint. For DSR-Bench, all question-answer pairs are further refined by human annotators to ensure accuracy.

\subsection{Video Curation}
\label{sec:3.1}
We begin with Koala-36M \cite{wang2025koala}, a diverse in-the-wild video corpus originally curated for video generation.  It is a large-scale, general-purpose video dataset encompassing a wide range of scenes. Moreover, it has been preprocessed to remove low-quality videos and each video is accompanied by a caption describing objects and overall event. 
%Prof. Qi's Version
%We begin with Koala-36M \xjqi{add citations}, a diverse in-the-wild video corpus originally curated for video generation.  It is a large-scale, general-purpose video dataset encompassing a wide range of scenes. Moreover, it has been preprocessed to remove low-quality videos, and each video is accompanied by a caption describing objects and overall event. 

However, many videos in Koala-36M are unsuitable for DSR, as they exhibit no or minimal object motion--such as articulated movements--without significant object position change.  To ensure meaningful spatial content, we filter out clips with negligible global or object motion as in Stage 1 of Figure \ref{img:2}: (i) For DSR-Train, we use DeepSeek-R1 \cite{guo2025deepseek} to filter candidates based on captions; (ii) For DSR-Bench, we adopt Gemini-2.5-Pro \cite{comanici2025gemini}, prompting directly on video content for more reliable selection. We retain videos between 20s–120s to balance temporal context and computational cost. After filtering, we obtain 10,000 videos for training and a curated subset of 575 videos for evaluation. This step ensures scene diversity, real-world dynamics, and sufficient temporal span for multi-object spatial understanding. 

\subsection{Geometric Clue Extraction} 
\label{sec:3.2}
%\xjqi{You need a figure for this}
If without awareness to absolute scale, vision foundation models cannot produce reliable metric-scale 3D structure. Worse still, annotating ground-truth scale is labor-intensive and often inconsistent, resulting in inaccurate question-answer pairs (QA). Therefore, with the curated videos in hand, we primarily process the monocular footage with vision foundation models that yield relative (non-metric) 3D structure to enable generation of qualitative, trend-based QAs (e.g., larger/smaller, left/right, faster/slower) for faithful supervision and evaluation. For DSR-Train, we uniformly sample 32 frames per video to control computation; for DSR-Bench, we sample at 1 FPS to maximize answer fidelity. Because objects may enter or exit, we randomly select a time sub-interval per QA instance and keep only the frames within it, then apply vision foundation models to extract the geometric cues necessary for QA construction.
\begin{table}[t]
\caption{Type, target object and description of candidate questions.}
\centering
\renewcommand{\arraystretch}{1.3}
\resizebox{0.71\width}{!}{
\begin{tabular}{ccc}
\shline
Type                                                           & Target Object                                                                              & Description                                                                                                           \\ \shline
Distance                                                       & \begin{tabular}[c]{@{}c@{}}Two random objects\\ (agents or non-agents)\end{tabular} & \begin{tabular}[c]{@{}c@{}}Determining how the distance between\\ target objects changes over time\end{tabular}       \\ \hline
Direction                                                      & \begin{tabular}[c]{@{}c@{}}Two random objects\\ (agents or non-agents)\end{tabular} & \begin{tabular}[c]{@{}c@{}}Identifying how the direction of one\\ object with respect to another changes\end{tabular} \\ \hline
Orientation                                                    & \begin{tabular}[c]{@{}c@{}}One random object\\ (agent only)\end{tabular}            & \begin{tabular}[c]{@{}c@{}}Describing how the orientation \\ of an agent evolves\end{tabular}                        \\ \hline
Speed                                                          & \begin{tabular}[c]{@{}c@{}}One random object\\ (agent or non-agent)\end{tabular}    & \begin{tabular}[c]{@{}c@{}}Assessing how the speed\\ of an object varies\end{tabular}                                 \\ \hline
\begin{tabular}[c]{@{}c@{}}Speed\\ Comparison\end{tabular}     & \begin{tabular}[c]{@{}c@{}}Two random objects\\ (agents or non-agents)\end{tabular} & \begin{tabular}[c]{@{}c@{}}Comparing the speed of two target \\ objects throughout the video\end{tabular}             \\ \hline
\begin{tabular}[c]{@{}c@{}}Direction\\ Prediction\end{tabular} & \begin{tabular}[c]{@{}c@{}}One random object\\ (agent or non-agent)\end{tabular}    & \begin{tabular}[c]{@{}c@{}}Predicting the moving direction\\ of a target object at the end of video\end{tabular}      \\ \shline
\end{tabular}
}
\label{tab:1}
\end{table}
As shown in Stage 2 of Figure \ref{img:2}, at the scene level we estimate camera poses and local point clouds using $\pi^3$ \cite{wang2025pi}, which provides robust relative-scale reconstructions in dynamic, in-the-wild settings. At the object level, we use DeepSeek-R1 (caption-guided) to extract agent \& non-agent categories and Grounded SAM2 \cite{ren2024grounded} for tracking and segmentation, yielding temporally consistent masks. Each mask is lifted to 3D by projecting it onto the corresponding point cloud from the $\pi^3$ model \cite{wang2025pi}. The mean of the associated points serves as the object’s 3D center at each timestamp, producing smooth 3D trajectories across the sub-interval. For objects of agent classes, we further estimate orientations (azimuth, elevation, roll w.r.t. the camera) with Orient Anything \cite{wangorient} while orientations for non-agents  are omitted to avoid spurious estimates. We prune objects that are not continuously visible to ensure reliability. The resulting geometric cues, including camera poses, per-timestamp geometry, multi-object trajectories, 3D centers and agent orientations, form a compact, expressive basis for constructing questions and answers for DSR task.

\begin{table}[t]
\caption{Basic answer choices for each type of question.}
\centering
\renewcommand{\arraystretch}{1.3}
\resizebox{0.85\width}{!}{
\begin{tabular}{cc}
\shline
Type                                                                                         & Basic Choices                                                                                                                                                                                                   \\ \shline
\begin{tabular}[c]{@{}c@{}}Distance \&\\ Speed\end{tabular}                                   & \begin{tabular}[c]{@{}c@{}}(1) Keep nearly constant then become larger;\\ (2)Keep nearly constant then become smaller; \\ (3) Keep nearly constant; (4) Become larger; \\ (5) Become smaller\end{tabular} \\ \hline
\begin{tabular}[c]{@{}c@{}}Direction \&\\ Orientation \&\\ Direction\\ Prediction\end{tabular} & \begin{tabular}[c]{@{}c@{}}Combinations of (1) Front/Behind; \\ (2) Left/Right; (3) Above/Below\end{tabular}                                                                                              \\ \hline
\begin{tabular}[c]{@{}c@{}}Speed\\ Comparison\end{tabular}                                   & \begin{tabular}[c]{@{}c@{}}(1) Nearly the same; (2) The former is faster; \\ (3) The latter is fater\end{tabular}                                                                                         \\ \shline
\end{tabular}
}
\label{tab:2}
\end{table}

\begin{figure*}[t]
\centering
\includegraphics[width=\linewidth]{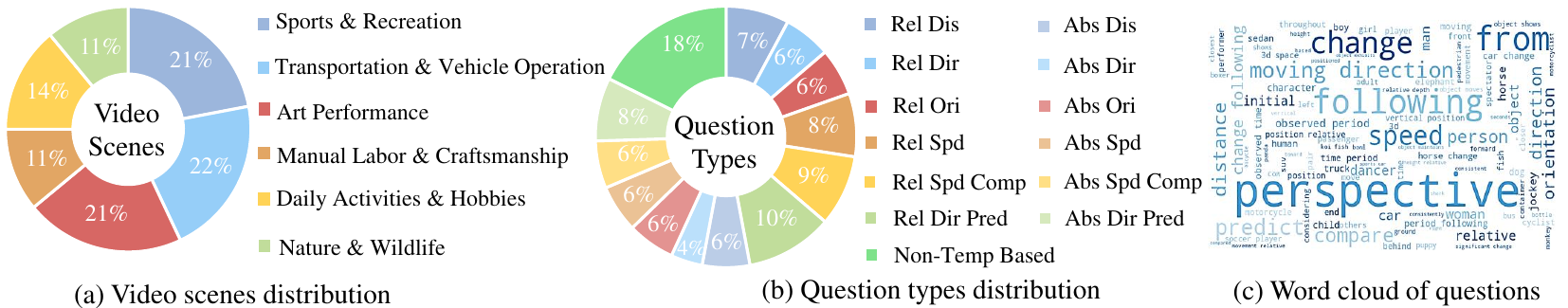}
\caption{Statistical overview of our DSR-Bench. In (a), we present the proportion of videos corresponding to different scene classes. In (b), we show the proportion of questions of various types. In (c), we depict a word cloud of the questions.}
\label{img:3}
\end{figure*} 

\subsection{Data Generation}
\label{sec:3.3}
\begin{comment}
With all required information collected, we generate question-answer pairs, categorized into template-based ones for basic dynamic spatial reasoning ability and non-template-based ones for overall ability. Each template-based question consists of four key components, which are time period, question type, target objects and observing viewpoint. The time period is identical to sub-interval determined before, while question type and target objects are randomly chosen from the following options:
\end{comment}

\noindent \textbf{Types of data.} With all geometric cues in place, we synthesize multiple-choice QA pairs in two families: template-based items that probe core dynamic-spatial skills and non-template (free-form) items that assess holistic reasoning.  Each template QA is defined over a time sub-interval, a question type, a set of target objects, and an observing viewpoint. We instantiate six types of template QA whose target objects and descriptions are listed in Table \ref{tab:1}.

\noindent\textbf{Viewpoints.} Spatial intelligence is inherently tied to the \emph{observer’s} viewpoint, which defines the operative coordinate system. To encourage reasoning across different situations, we vary the viewpoint and its mobility: (i) viewpoint can be posed from the \emph{camera} or from an \emph{agent}; (ii) viewpoint can follow the motion of its corresponding agent and evolve across time sub-interval (relative viewpoint) or fixed to its status at one specific timestamp (absolute viewpoint). We transform all object 3D centers into the selected reference coordinate using camera poses and agent orientations, enabling questions that require \emph{egocentric-allocentric} transformation reasoning rather than passive observation. 

%Prof. Qi's Version
%\xjqi{Please check the following as I cannot understanding your writing, just based on my guess} 

%\noindent\textbf{Viewpoints.} Spatial intelligence is inherently tied to the \emph{observer’s} viewpoint, which defines the operative coordinate frame. To encourage reasoning across reference frames, we vary the observer: questions may be posed from the \emph{camera} (absolute viewpoint) or from a \emph{moving agent} (relative viewpoint). We transform all object 3D centers into the selected reference frame using camera poses and agent orientations, enabling questions that require \emph{egocentric-- allocentric} transformation reasoning rather than passive observation. 

\begin{comment}
Regarding observing viewpoint, since egocentric-allocentric transformation is a crucial aspect of spatial reasoning as discussed in [], we randomly select one agent as observer rather than restricting it to camera. Moreover, since one object's motion will be a relative one if the viewpoint is moving as well while absolute when the viewpoint is static, we categorize viewpoint into absolute where the viewpoint remains fixed, and relative, where the viewpoint follows the observer movement, and select one of them when generating questions. Objects' 3D centers are transformed to the coordinates under the selected viewpoint according to camera poses and observer orientations. 
\end{comment}

\noindent \textbf{Question-answer pair generation.} 
When generating one template-based QA, we randomly select one question type, required target objects according to Table \ref{tab:1} and one viewpoint as well as its mobility. Together with time sub-interval selected before, we construct the question according to corresponding template. For multiple choices, since monocular 3D reconstruction produces relative-scale geometry and humans also struggle to consistently judge metric values in unconstrained videos, our choices are qualitative rather than numeric and are combinations of basic choices in Table \ref{tab:2}. We generate the correct answer by comparing the queried attribute every 2 adjacent frames to derive one basic choice and then merging consecutive identical states to form a concise, procedural answer that reflects the state evolvement over time, rather than a single frame snapshot.

While template-based QAs mainly evaluate fundamental spatial-temporal abilities, they may not fully reflect how humans naturally query motion and geometry. To broaden linguistic variety and reasoning patterns, we additionally employ DeepSeek-R1 to auto-generate non-template-based questions and answers conditioned on extracted 3D trajectories, object identities and viewpoints, aiming to probe more general, free-form understanding of scene dynamics.

%Prof. Qi's Version
%While template-based QA pairs evaluate fundamental spatial-temporal abilities, they may not fully reflect how humans naturally query motion and geometry. To broaden linguistic variety and reasoning patterns, we additionally employ DeepSeek-R1 to auto-generate non-template questions and answers. These questions are conditioned on recovered 3D trajectories, object identities, and viewpoints, and aim to probe more general, free-form understanding of scene dynamics. \xjqi{All prompts and generation details are described in the Appendix.}

%These template-based question-answer pairs are primarily related to basic abilities and not general enough for training and evaluting model's overall capability. Therefore, we additionally prompt DeepSeek-R1 to generate non-template-based questions and answers according to 3D object trajectories from selected viewpoint. The prompt is provided in the Appendix.

Together, this hybrid QA generation strategy yields a data suite that rigorously targets dynamic spatial reasoning: it grounds every question in 4D evidence, supports both egocentric and allocentric querying, and provides fine-grained procedural answers that evaluate a model's understanding of continuous object motion, spatial change, and multi-object interactions. All prompts, templates and generation details are described in the Appendix.

\subsection{Benchmark Statistics and Comparisons}
\label{sec:3.4}
\begin{table*}[t]
\caption{Comparison between our DSR-Bench and prior benchmarks from different aspects.}
\centering
\renewcommand{\arraystretch}{1.3}
\resizebox{0.99\linewidth}{!}{
\begin{tabular}{ccccccc}
\shline
Benchmark                           & Input                                 & Scene   & \begin{tabular}[c]{@{}c@{}}Target Object\\Number\end{tabular}  & \begin{tabular}[c]{@{}c@{}}Viewpoint\\Transformation\end{tabular} & \begin{tabular}[c]{@{}c@{}}3D Awareness\\ Requirement\end{tabular} & \begin{tabular}[c]{@{}c@{}}Answer Temporal\\ Granularity\end{tabular}\\ \shline
MMSI-Bench \cite{yang2025mmsi}                      & \multirow{3}{*}{Two Images}                     &        In-the-Wild            &         Multi-Object             &              $\checkmark$              &              \multirow{3}{*}{-}    &      \multirow{3}{*}{-}        \\
SpatialScore \cite{wu2025spatialscore}                     &                                       &            In-the-Wild           &         Single-Object             &             $\times$               &                    &           \\
SAT \cite{ray2024sat}                              &                                       &           In-the-Wild           &          Single-Object    &             $\times$               &            &                   \\ \hline
\multicolumn{1}{l}{DynSuperCLEVR \cite{wangcompositional}} & \multicolumn{1}{c}{\multirow{5}{*}{Video}} & \multicolumn{1}{c}{Driving} & \multicolumn{1}{c}{Multi-Object} & \multicolumn{1}{c}{$\times$}       & \multicolumn{1}{c}{Medium}    & Coarse      \\
VLM4D \cite{zhou2025vlm4d}                            & \multicolumn{1}{c}{}                  &          In-the-Wild            &       Single-Object               &             $\times$               &          Weak        & Coarse             \\
STI-Bench \cite{li2025sti}                        & \multicolumn{1}{c}{}                  &     Driving            &           Single-Object           &             $\times$               &            Medium        & Coarse           \\
OmniSpatial \cite{jia2025omnispatial}                      & \multicolumn{1}{c}{}                  &    Human-Object Interaction           &            Single-Object         &              $\times$              &                Weak     & Coarse      \\
DSR-Bench                       & \multicolumn{1}{c}{}                  &    In-the-Wild           &            Multi-Object         &              $\checkmark$              &                Strong     & Fine      \\
\shline
\end{tabular}
}
\label{tab:3}
\end{table*}

We first report the comprehensive statistics of \textbf{DSR-Bench}. Figure \ref{img:3}(a), (b)  summarizes the distributions of videos and questions belonging to different categories. Following Gemini-2.5-Pro's advice, videos span six broad categories: {Sports \& Recreation};  {Transportation \& Vehicle Operation}; {Art Performance}; {Manual Labor \& Craftsmanship};  {Daily Activities \& Hobbies}; {Nature \& Wildlife}. There are totally 1484 questions organized into 12 template-based types (the Cartesian product of viewpoint mobility and question type) plus one non-template class. The results show that both the video and question distributions are balanced, supporting broad generalization. We also provide the word cloud of questions in Figure \ref{img:3}(c) to show our evaluation emphasis on the complete state evolvement procedure, reasoning from different viewpoints and multi-object interactions.

%Prof. Qi's Version
%We report comprehensive statistics for \textbf{DSR-Bench} and situate it among prior spatial-reasoning benchmarks. \xjqi{Figure}  summarizes the distribution of videos and questions. Following the Gemini-2.5-Pro taxonomy, videos span six broad categories: {Sports \& Recreation};  {Transportation \& Vehicle Operation}; {Art Performance}; {Manual Labor \& Craftsmanship};  {Daily Activities \& Hobbies}, and {Nature \& Wildlife}. Questions are organized into 12 template-based types (the Cartesian product of viewpoint and question type) plus one non-template class. As shown in \xjqi{Figure x} both the video and question distributions are well balanced, supporting broad generalization. 

\begin{comment}
We provide the statistics of our DSR-Bench and compare it with previous spatial reasoning benchmarks. Figure () presents the proportions of videos and questions across different classes. Following the categorization of Gemini-2.5-Pro, the videos are divided into 6 classes: (1) Sports \& Recreation; (2) Transportation \& Vehicle Operation; (3) Art Performance; (4) Manual Labor \& Craftsmanship; (5) Daily Activities \& Hobbies; (6) Nature \& Wildlife. Questions are categorized into 12 template-based types according to the combination of viewpoint and question type, plus 1 non-template-based class. As shown in Figure (), both video and question distributions are well balanced in DSR-Bench.
\end{comment}

Second, we contrast DSR-Bench with prior spatial reasoning benchmarks containing object position changes in Table \ref{tab:3}. Part of benchmarks only consider reasoning the difference between two images, prevented from long-term evaluation. For benchmarks with video inputs, while most of them draw from domain-constrained sources (e.g., autonomous driving \cite{sun2020scalability,caesar2020nuscenes} or human-object interaction \cite{liu2022hoi4d}), DSR-Bench is built from \emph{in-the-wild} videos, enabling evaluation in diverse and unconstrained environments. Except for DynSuperCLEVR \cite{wangcompositional}, earlier efforts largely emphasize single-object motion from the camera’s viewpoint, limiting assessment of multi-object and multi-viewpoint reasoning. 
%Prof. Qi's Version
%Table \ref{tab:3} contrasts DSR-Bench with existing datasets. Whereas most prior benchmarks draw from domain-constrained sources (e.g., autonomous driving \xjqi{add citations} or human-object interaction \xjqi{add citations}), DSR-Bench is built from \emph{in-the-wild} videos, enabling evaluation in diverse, unconstrained environments. Except for DynSuperCLEVR \xjqi{add citations}, earlier efforts largely emphasize single-object motion from the camera’s viewpoint, limiting assessment of multi-object and multi-viewpoint reasoning. 

Furthermore, we quantify 3D knowledge requirements in order to resolve a task. We conduct two complementary checks: (i) \emph{object-level}, which uses DeepSeek-R1 to judge whether a question does not require 3D attributes such as orientation, shape, or size (a benchmark is marked as requiring object-level knowledge if less than half of responses are positive); and (ii) \emph{scene-level}, which uses DeepSeek-R1 to determine whether a question can be answered from 2D changes alone (if fewer than half respond ``yes'' scene-level knowledge is deemed necessary). We then label overall 3D demand as \emph{weak}, \emph{medium}, or \emph{strong} depending on whether none, one, or both knowledge types are required. Our results show that proportion of positive answers about two checks for different benchmarks are: (1) DynSuperCLEVR--63\% \& 24\%; (2) VLM4D--73\% \& 85\%; (3) STI-Bench--69\% \& 21\%; (4) OmniSpatial--56\% \& 79\%; (5) DSR-Bench--34\% \& 18\%. Therefore, their 3D demand levels are medium, weak, medium, weak and strong.

Finally, while most benchmarks provide coarse, aggregate answers, DSR-Bench supplies \emph{fine-grained, procedural} answers that describe the change over time, compelling VLMs to reason about continuous dynamics rather than isolated outcomes. We also quantify this by prompting DeepSeek-R1 to judge whether an answer describes the changing procedure of the asked attribute across video. The proportion of positive answers for different benchmarks are: (1) DynSuperCLEVER--2\%; (2) VLM4D--19\%; (3) STI-Bench--22\%; (4) OmniSpatial--18\%; (5) DSR-Bench--78\%. Therefore, aside from our DSR-Bench, all previous benchmarks are dominated by course-grained answers. Collectively, these properties make DSR-Bench a comprehensive benchmark for DSR evaluation. 

\begin{table*}[ht]
\caption{Performance comparison among different VLMs on different subtasks of DSR-Bench.}
\centering
\renewcommand{\arraystretch}{1.3}
\resizebox{0.76\width}{!}{
\begin{tabular}{ccccccccccccccc}
\shline
\multicolumn{1}{c|}{\multirow{2}{*}{Models}}   & \multicolumn{14}{c}{Subtask Types}                                                                                                                                                                                                                                                                                                                                                                                                                                                                                                                                                                                                                                                                                                              \\ \cline{2-15} 
\multicolumn{1}{c|}{}                          & \begin{tabular}[c]{@{}c@{}}Abs\\ Dis\end{tabular} & \begin{tabular}[c]{@{}c@{}}Abs\\ Dir\end{tabular} & \begin{tabular}[c]{@{}c@{}}Abs\\ Ori\end{tabular} & \begin{tabular}[c]{@{}c@{}}Abs\\ Spd\end{tabular} & \begin{tabular}[c]{@{}c@{}}Abs Spd\\ Comp\end{tabular} & \begin{tabular}[c]{@{}c@{}}Abs Dir\\ Pred\end{tabular} & \begin{tabular}[c]{@{}c@{}}Rel\\ Dis\end{tabular} & \begin{tabular}[c]{@{}c@{}}Rel\\ Dir\end{tabular} & \begin{tabular}[c]{@{}c@{}}Rel\\ Ori\end{tabular} & \begin{tabular}[c]{@{}c@{}}Rel\\  Spd\end{tabular} & \begin{tabular}[c]{@{}c@{}}Rel Spd\\ Comp\end{tabular} & \begin{tabular}[c]{@{}c@{}}Rel Dir\\ Pred\end{tabular} & \begin{tabular}[c]{@{}c@{}}Non-Temp\\ Based\end{tabular} & Avg \\ \shline
\rowcolor{black!10}\multicolumn{15}{c}{\textit{Propriety Models}}                                                                                                                                                                                                                                                                                                                                                                                                                                                                                                                                                                                                                                                                                                                                                   \\ \shline
\multicolumn{1}{c|}{GPT-4o \cite{hurst2024gpt}}                    &          18.8                                          &             29.2                                       &                  26.8                                  &        29.7                                            &          21.5                                          &                    26.2                                &                          24.1                          &                23.8                                    &                   17.2                                      &                    32.3                                     &                                22.8                         &             24.4                                            &            34.7                                               &   26.4  \\
\multicolumn{1}{c|}{GPT-5 \cite{openai2025gpt5}}                     &            21.1                                        &                  41.5                                  &                       48.7                             &           34.5                                         &            33.3                                        &                      34.7                              &                               17.2                     &              44.3                                      &                    41.9                                     &                       21.2                                  &                            25.0                             &            30.9                                             &                  26.7                                         &    30.8 \\
\multicolumn{1}{c|}{Gemini-2.5-Flash \cite{comanici2025gemini}}          &              18.8                                      &                    27.6                                &                         19.5                           &          25.0                                          &               23.6                                     &                        22.0                            &                           11.2                         &               28.4                                     &                  30.8                                       &                     23.2                                    &                               17.1                          &          22.6                                               &                  38.8                                         &  24.9   \\
\multicolumn{1}{c|}{Gemini-2.5-Pro \cite{comanici2025gemini}}            &              20.0                                      &                     44.6                               &                       53.6                             &        27.3                                            &              38.7                                      &                          30.5                          &                            23.2                        &              32.9                                      &                     43.2                                    &                    17.1                                     &                                 28.5                        &             27.9                                            &              34.3                                             &   31.7  \\ \shline
\rowcolor{black!10}\multicolumn{15}{c}{\textit{Video Understanding Models}}                                                                                                                                                                                                                                                                                                                                                                                                                                                                                                                                                                                                                                                                                                                                         \\ \shline
\multicolumn{1}{c|}{LLaVA-Video-7B \cite{zhang2024video}}            &                22.3                                    &                     16.9                               &                      25.6                              &          33.3                                          &                    45.1                                &                        24.5                            &                           24.1                         &               15.9                                     &                   17.2                                      &                          19.1                               &                          24.2                               &            21.4                                             &             33.9                                              &  25.9   \\
\multicolumn{1}{c|}{VideoRefer \cite{yuan2025videorefer}}                &            23.5                                        &                    18.4                                &                          25.6                          &        33.5                                            &                  45.4                                  &                         27.1                           &                                25.0                    &                   16.1                                 &                      18.5                                   &                          20.2                               &                               26.4                          &          22.6                                               &           34.7                                                &  26.9   \\
\multicolumn{1}{c|}{LongVILA-R1 \cite{chen2025scaling}}               &             20.0                                       &              21.5                                      &                           23.1                         &           21.4                                         &                 37.6                                   &                        22.8                            &                         24.1                           &              26.1                                      &                 28.3                                        &                      22.2                                   &                                  17.8                       &            20.8                                             &          33.9                                                 &   25.3  \\ \shline
\rowcolor{black!10}\multicolumn{15}{c}{\textit{General-Purpose Models}}                                                                                                                                                                                                                                                                                                                                                                                                                                                                                                                                                                                                                                                                                                                                             \\ \shline
\multicolumn{1}{c|}{Qwen2.5-VL-7B \cite{bai2025qwen2}}             &                 18.8                                   &                      15.3                              &                         14.6                           &           42.8                                         &                  29.0                                  &                           19.4                        &                          31.8                          &                   19.3                                 &                       11.1                                  &                        22.2                                 &                             19.2                            &            20.2                                             &                30.1                                           &   23.5  \\
\multicolumn{1}{c|}{Qwen2.5-VL-32B \cite{bai2025qwen2}}            &               31.7                                     &                      21.5                              &                         23.1                           &        44.0                                            &                    36.5                                &                           25.4                         &                            27.5                        &             23.8                                       &                 37.0                                        &                           27.2                              &                              29.2                           &             21.4                                            &               36.2                                            &   29.9  \\
\multicolumn{1}{c|}{Qwen3-VL-8B-Instruct \cite{qwen2025qwen3}}      &                 23.5                                   &                     24.6                               &                           42.6                         &        29.7                                            &                 27.9                                   &                        33.8                            &                             18.1                       &              28.4                                      &                     34.5                                    &                            24.2                             &                             22.1                            &            27.9                                             &              33.5                                             &   28.7  \\
\multicolumn{1}{c|}{Qwen3-VL-30B-A3B-Instruct \cite{qwen2025qwen3}} &                 25.8                                   &                      27.6                              &                        46.3                            &           30.9                                         &                  31.1                                  &                           34.7                         &                            20.6                        &                29.5                                    &                      37.0                                   &                         28.2                                &                              24.2                           &            31.5                                             &             35.4                                              &   31.1  \\
\multicolumn{1}{c|}{InternVL3.5-8B \cite{wang2025internvl3}}            &               23.5                                     &                  27.6                                  &                           28.0                         &           34.5                                         &                    24.7                                &                          27.9                          &                              22.4                      &             17.0                                       &                  19.7                                       &                            28.2                             &                               30.0                          &            14.2                                             &             30.1                                              &   25.4  \\
\multicolumn{1}{c|}{InternVL3.5-38B \cite{wang2025internvl3}}           &               25.8                                     &                  27.8                                  &                         29.2                           &            34.2                                        &                 24.7                                   &                          28.5                          &                          26.7                          &            16.3                                        &                           23.4                              &                       29.2                                  &                                   32.1                      &             15.4                                            &                     31.3                                      &   26.7  \\ \shline
\rowcolor{black!10}\multicolumn{15}{c}{\textit{Spatial Reasoning Models}}                                                                                                                                                                                                                                                                                                                                                                                                                                                                                                                                                                                                                                                                                                                                           \\ \shline
\multicolumn{1}{c|}{VLM-3R \cite{fan2025vlm}}                     &                 28.2                                   &                 27.6                                   &                          31.7                          &            42.8                                        &                    38.7                                &                        33.0                            &                              34.4                      &              23.8                                      &                        30.8                                 &                          22.2                               &                        26.4                                 &            29.1                                             &           35.0                                                &   31.4  \\
\multicolumn{1}{c|}{VG-LLM \cite{zheng2025learning}}                    &                  55.2                                  &                   32.3                                 &                           58.5                         &          57.1                                          &                       51.6                             &                           32.2                         &                            56.0                        &             36.3                                       &                   32.0                                      &                          30.3                               &                                  32.1                       &            29.1                                             &                 27.9                                          & 38.4\\
\multicolumn{1}{c|}{\textbf{Ours}}                      &                 \textbf{87.0}                                   &                  \textbf{73.8}                                  &                          \textbf{84.1}                          &         \textbf{73.8}                                           &                    \textbf{72.0}                                &                        \textbf{35.5}                            &                              \textbf{75.8}                      &            \textbf{76.1}                                        &                  \textbf{77.7}                                       &                       \textbf{60.6}                                  &                                   \textbf{37.1}                      &            \textbf{35.1}                                             & \textbf{46.4} & \textbf{58.9}                                            \\ \shline
\end{tabular}
}
\label{tab:4}
\end{table*}
\section{Geometric Prior Enhanced VLM for Dynamic Spatial Reasoning}
\label{sec:gate}

\begin{figure}[t]
\centering
\includegraphics[width=0.97\linewidth]{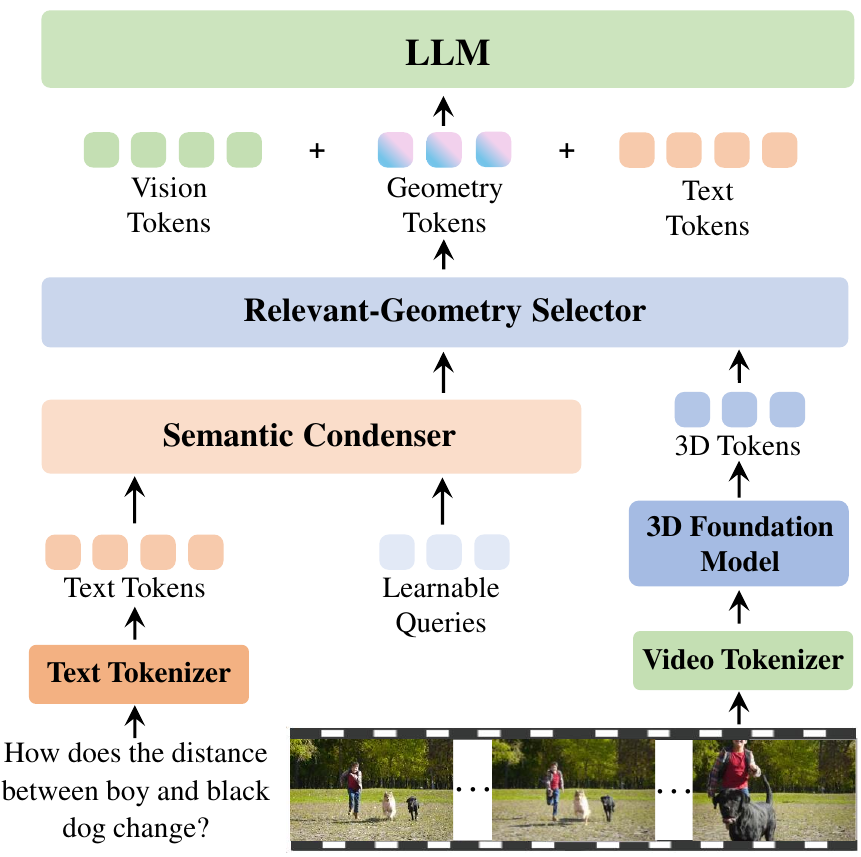}
\caption{Illustraction of our proposed GSM that consists of two stacked Q-Formers. The first Q-Former condenses question semantics, and the second one extracts question-relevant geometric knowledge into a compact set of geometry tokens. These tokens are appended to original vision tokens to be processed by LLM.}
\label{img:4}
\end{figure} 

A common strategy to enhance spatial reasoning is to inject features from pretrained 3D/4D foundation models into VLMs. Prior works typically either (i) apply cross-attention between vision tokens and 3D tokens (e.g., from CUT3R \cite{wang2025continuous}), or (ii) directly add geometry features (e.g., from VGGT \cite{wang2025vggt}) to vision tokens \cite{fan2025vlm,zheng2025learning}. While these approaches improve spatial reasoning performance, they tend to degrade general video understanding (e.g., Video-MME, see Table \ref{tab:5} in Section \ref{sec:abl_1}). This happens because geometric models cam produce noisy geometric cues when processing in-the-wild videos, overwhelming the VLM and causing task-specific overfitting and reduced generality. 

%Prof. Qi's Version
%A common strategy to improve spatial reasoning in video VLMs is to inject features from pretrained 3D/4D foundation models. Prior works typically either (i) apply cross-attention between VLM vision tokens and geometry tokens (e.g., from CUT3R) \xjqi{add citations}, or (ii) directly concatenate geometry features (e.g., from VGGT) with vision tokens \xjqi{add citations}. While these approaches often boost performance on spatial-reasoning benchmarks, they tend to degrade general video understanding (e.g., Video-MME) \xjqi{add evidance, refer to table....}. This happens because geometry models processing in-the-wild videos may produce irrelevant or noisy geometric cues, overwhelming the VLM and causing task-specific overfitting and reduced generality. 

To address this trade-off, we propose a lightweight Geometry Selection Module (\textbf{GSM}) to selectively incorporate geometric priors into VLMs. Instead of exposing VLM to all 3D tokens, GSM retrieve a compact, task-relevant subset of geometric knowledge based on questions. Concretely, as in Figure \ref{img:4}, given a video, we compute (i) the VLM vision tokens $\mathbf{T}_{\mathrm{vis}}$, (ii) the question tokens $\mathbf{T}_{\mathrm{text}}$, (iii) 3D tokens $\mathbf{T}_{\mathrm{3D}}$ obtained by applying $\pi^3$ encoder \cite{wang2025pi} to video frames to form the geometric prior. GSM employs \emph{two stacked Q-Formers} to produce a fixed-size set of $N$ geometry tokens:

\begin{enumerate}
  \item \textbf{Language condensation.} The first Q-Former (Semantic Condenser) takes $N$ learnable queries and attends to $\mathbf{T}_{\mathrm{text}}$, distilling the question semantics into a set of language-conditioned query embeddings $\mathbf{Q}_{\mathrm{lang}} \!\in\! \mathbb{R}^{N \times d}$.
  \item \textbf{Selective geometry retrieval.} The second Q-Former (Relevant-Geometry Selector) attends $\mathbf{Q}_{\mathrm{lang}}$ to $\mathbf{T}_{\mathrm{3D}}$, extracting \emph{only the geometry relevant to the question} and yielding compact geometry tokens $\mathbf{Q}_{\mathrm{geo}} \!\in\! \mathbb{R}^{N \times d}$.
\end{enumerate}

%Prof. Qi's Version
%\begin{enumerate}
%  \item \textbf{Language condensation.} A first Q-Former takes $N$ learnable queries and attends to $\mathbf{T}_{\mathrm{text}}$, distilling the question semantics into a set of language-conditioned query embeddings $\mathbf{Q}_{\mathrm{lang}} \!\in\! \mathbb{R}^{N \times d}$.
%  \item \textbf{Selective 4D retrieval.} A second Q-Former uses $\mathbf{Q}_{\mathrm{lang}}$ to attend over $\mathbf{T}_{\mathrm{4D}}$, extracting \emph{only the geometry relevant to the question} and yielding compact \emph{geometric} queries $\mathbf{Q}_{\mathrm{geo}} \!\in\! \mathbb{R}^{N \times d}$.
%\end{enumerate}
Because $N$ is fixed, GSM presents the language model with a \emph{bounded} and \emph{task-aligned} geometric summary, avoiding the brittleness of directly exposing long, variable-length $\mathbf{T}_{\mathrm{3D}}$ to the VLM.
%\paragraph{Fusion with vision tokens.}
The extracted geometry tokens are concatenated with vision tokens and question tokens:
\vspace{-4pt}

\[
\tilde{\mathbf{T}}_{\mathrm{total}} = [\, \mathbf{T}_{\mathrm{vis}} \,;\, \mathbf{Q}_{\mathrm{geo}} \,;\, \mathbf{T}_{\mathrm{text}} \,],
\]
and the combined stream is passed to the LLM head. This late, compact fusion injects essential geometric priors while preserving the VLM’s general reasoning capacity.

\begin{comment}
Concretely, to obtain 4D prior, we apply the encoder of $\pi^3$ to each video, producing a sequence of 3D tokens $\mathcal{T}_\text{4D}$. A straightforward approach to extract question-relevant 3D information would be to perform cross-attention between the question tokens $\mathcal{T}_\text{text}$ and $\mathcal{T}_\text{3D}$. However, since the length of $\mathcal{T}_\text{text}$ varies across different questions, such a design may generalize poorly. To address this, we introduce a two-stage Q-Former structure that generates a fixed number ($N$) of tokens. Specifically, the first Q-Former takes $N$ learnable queries and $\mathcal{T}_\text{text}$ as input, condensing the question information into the queries. The second Q-Former then refines these queries by attending to $\mathcal{T}\text{3D}$, extracting and encoding the question-relevant 3D knowledge. As a result, the refined query tokens encapsulate compact and task-specific 3D representations with a consistent size.

Finally, we concatenate these query tokens with the vision tokens from the VLM and use the combined representation as input to the subsequent language model. The overall training objective remains the standard cross-entropy loss between the model’s output logits and the ground-truth answer tokens. 

\end{comment}
\noindent\textbf{Discussions.} GSM is architecture-agnostic (works with different video VLM backbones and geometry encoders), parameter-efficient (fixed $N$ queries), and robust to question length (language condensation normalizes variable $\mathbf{T}_{\mathrm{text}}$). Empirically, it yields strong gains for DSR while maintaining performance on general video/VLM benchmarks.

\section{Experiments}

\subsection{Experimental Settings}

We adopt Qwen2.5-VL-7B as our base model and integrate it with the proposed GSM. The model is trained on 50K question-answer pairs from DSR-Train. All components are trainable except freezing the vision encoder of Qwen2.5-VL-7B. The number of learnable queries $N$ in GSM is set to 32. The model is trained for 1 epoch the learning rate of 2$\times$10$^{-7}$ and the batch size of 32.

\subsection{Comparison with State-of-The-Arts}

We first compare our model’s performance against the following state-of-the-art VLMs on DSR-Bench and report their performance on different subtasks:
\begin{itemize}
\item Proprietary models, including GPT-4o \cite{hurst2024gpt}, GPT-5 \cite{openai2025gpt5}, Gemini-2.5-Flash \cite{comanici2025gemini}, Gemini-2.5-Pro \cite{comanici2025gemini};
\item Video understanding models, including LLaVA-Video-7B \cite{zhang2024video}, VideoRefer \cite{yuan2025videorefer}, LongVILA-R1 \cite{chen2025scaling};
\item General-purpose models, including Qwen2.5-VL-7B \cite{bai2025qwen2}, Qwen2.5-VL-32B \cite{bai2025qwen2}, Qwen3-VL-8B-Instruct \cite{qwen2025qwen3}, Qwen3-VL-30B-A3B-Instruct \cite{qwen2025qwen3}, InternVL3.5-8B \cite{wang2025internvl3}, InternVL3.5-38B \cite{wang2025internvl3};
\item Spatial reasoning models--VLM-3R \cite{fan2025vlm}, VG-LLM \cite{zheng2025learning}.
\end{itemize}

The results in Table \ref{tab:4} indicate that our model achieves the best performance across all subtasks, as well as the highest average performance. For models not explicitly designed for spatial reasoning, the performance is only marginally above random guess, revealing their limited capacity for this task. Even spatial reasoning models, though trained on static scenes, still fall short, underscoring the need of dedicated DSR training data. Together, these observations highlight the inherent challenges of DSR.

\subsection{Ablation Studies}
\label{sec:5.3}
\paragraph{Effect of GSM}
\label{sec:abl_1}
As discussed in Section \ref{sec:gate}, GSM enables integration of geometric prior from 3D foundation model into VLMs while mitigating performance degradation on general video understanding tasks. In this section, we compare GSM with the following training paradigms to demonstrate its effectiveness:
(1) SFT, which directly applies supervised fine-tuning to Qwen2.5-VL-7B on the same training dataset; and (2) Addition, which directly adds 3D tokens from $\pi^3$ to the vision tokens before training on the same dataset. For efficiency, we randomly sample 20K QAs from DSR-Train for training all models.

Furthermore, to evaluate on other 3D benchmarks and general video understanding tasks, we adopt VLM4D \cite{zhou2025vlm4d}, STI-Bench \cite{li2025sti} as additional spatial reasoning benchmarks and Video-MME \cite{fu2025video} as general video benchmark. The results in Table \ref{tab:5} show that training on DSR-Train helps improve performance not only on DSR-Bench, but also other benchmarks requiring dynamic spatial reasoning, regardless of training paradigm. Therefore, DSR-Train is effective for improving dynamic spatial reasoning capability. When compared with SFT, GSM helps improve performance on benchmarks requiring spatial reasoning, taking advantage of integration of geometric priors. When compared with Addition, GSM is competitive for dynamic spatial reasoning while retains general video understanding ability.
\begin{table}[t]
\caption{Comparison between GSW and other training methods.}
\centering
\renewcommand{\arraystretch}{1.3}
\resizebox{0.78\width}{!}{
\begin{tabular}{c|ccccc}
\shline
\multirow{2}{*}{Methods}                              & \multicolumn{5}{c}{Benchmarks}                   \\ \cline{2-6} 
                                                      & DSR-Bench & VLM4D & STI-Bench & Video-MME & Avg. \\ \shline
Baseline                                              &      23.5     &    43.1   &     33.2      &     60.2      &   40.0   \\
SFT                                                   &     54.4      &    46.7   &      34.6     &     60.1      &    48.9  \\
Addition &     57.7      &    48.5   &      35.3     &     48.6      &   47.5   \\
GSM                                                  &     57.4      &    48.3   &     35.2      &     59.9      &    50.2  \\ \shline
\end{tabular}
}
\label{tab:5}
\end{table}

\vspace{0.1in}\noindent\textbf{Effect of Query Number} 
\label{sec:abl_2}
In this section, we conduct experiments to evaluate the impact of learnable query number in GSM. Specifically, we vary the number of queries across 8, 16, 32, 64, training all models on the same dataset containing 20K QAs randomly sampled from DSR-Train.

The results in Table \ref{tab:6} show that the model with more queries performs better for dynamic spatial reasoning. However, the increased queries also harm general video understanding performance, resulting in lower average performance. Therefore, it is necessary to set a proper learnable query number to obtain the best overall performance.
\begin{table}[t]
\caption{Ablation of learnable query numbers.}
\centering
\renewcommand{\arraystretch}{1.3}
\resizebox{0.78\width}{!}{
\begin{tabular}{c|ccccc}
\shline
\multirow{2}{*}{\begin{tabular}[c]{@{}c@{}}Query\\ Number\end{tabular}} & \multicolumn{5}{c}{Benchmarks}                   \\ \cline{2-6} 
                                                                        & DSR-Bench & VLM4D & STI-Bench & Video-MME & Avg. \\ \shline
8                                                                &      55.7     &   47.2    &     34.6      &     59.9      &   49.3   \\
16                                                                     &    56.9       &    47.8   &     34.8      &     60.0      &    49.8  \\
32                   &      57.4     &   48.3    &     35.2      &     59.9      &   50.2   \\
64                                                                    &    57.6       &   48.5    &     35.1      &     59.2      &   50.0   \\ \shline
\end{tabular}
}
\label{tab:6}
\end{table}

\begin{figure}[t]
\centering
\setlength{\belowcaptionskip}{-2pt}
\includegraphics[width=\linewidth]{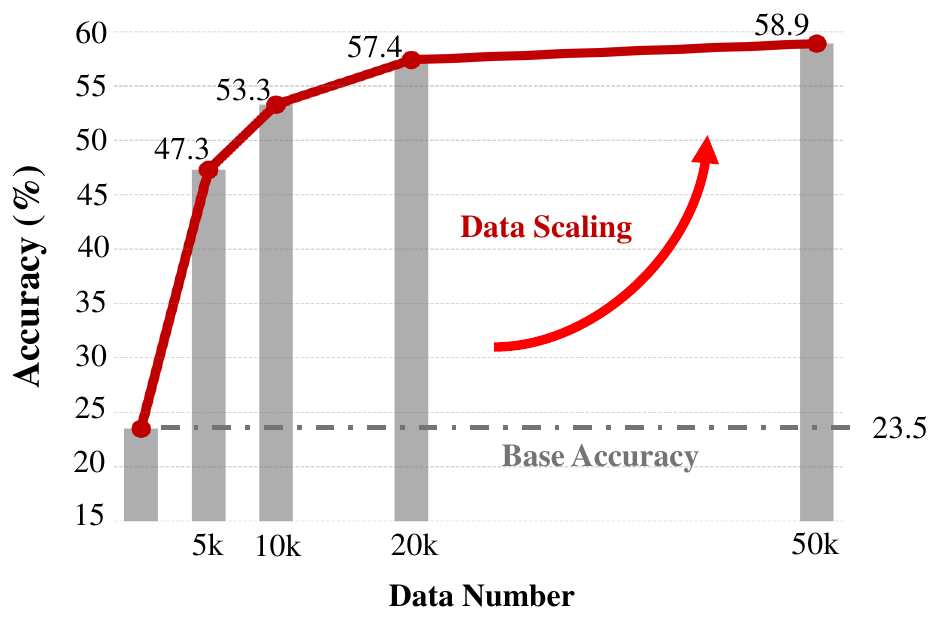}
\caption{Performance curve of accuracy on DSR-Bench with varying numbers of question-answer pairs for training.}
\label{img:5}
\end{figure}

\vspace{0.1in}\noindent\textbf{Scalability of DSR-Train} In this section, we analyze the scalability of DSR-Train by evaluating model performance when trained on varying numbers of QAs that are 5K, 10K, 20K and 50K. The results illustrated in Figure \ref{img:5} show that the model performs better on DSR-Bench when training data increases, showing the effectiveness of DSR-Train.
\section{Conclusion}
\label{sec:conclusion}
In this work, we present a unified framework DSR Suite for dynamic spatial reasoning in VLMs. It comprises an automated pipeline that builds DSR-Train for supervision and further human-refined DSR-Bench for comprehensive evaluation on in-the-wild videos. The data emphasize viewpoint transform, multi-object interact and fine-grained procedural answers. Beyond data, we introduced a lightweight Geometry Selection Module (GSM) that selectively integrates geometric priors via two stacked Q-Formers, avoiding noisy geometric overload to preserve general video understanding capability. Trained on DSR-Train, Qwen2.5-VL-7B + GSM achieves superior results on DSR-Bench without regressions on general benchmarks. We hope this suite and approach catalyze future work on 4D multimodal intelligence, including embodied perception, predictive reasoning and world modeling in dynamic environments.
\newpage
{
    \small
    \bibliographystyle{ieeenat_fullname}
    \bibliography{main}
}
\maketitlesupplementary
This supplementary material contains additional details of the main manuscript. Section \ref{sec:s1} presents additional implementation details for training data,  evaluation and QA generation. Section \ref{sec:s2} complements more experiments and analysis. Section \ref{sec:s3} illustrates some examples of QA in our DSR-Train and DSR-Bench.

\section{Implementation Details}
\label{sec:s1}
\subsection{Training Data Details}
\label{sec:s1.1}
\begin{figure}[ht]
\centering
\setlength{\belowcaptionskip}{-6pt}
\includegraphics[width=\linewidth]{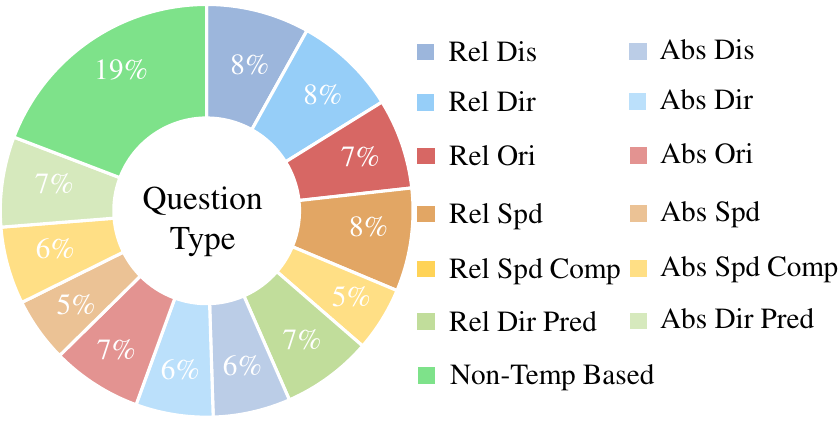}
\caption{Proportion of questions types in our DSR-Train.}
\label{img:s1}
\end{figure}

For 50K QAs in DSR-Train, we present the proportion of different question types in Figure \ref{img:s1}.

The illustration shows that DSR-Train has similar question type distribution with DSR-Bench as in Figure \ref{img:3}(b) of the main manuscript. For the randomly sampled 5K, 10K, 20K QAs for training in Section 5.3, we restrict their distributions to be the same as that in Figure \ref{img:s1}.

\subsection{Evaluation Details}
During evaluation, we uniformly sample 32 frames from each input video as visual inputs for all VLMs. For all models except Qwen series \cite{qwen2025qwen3, bai2025qwen2} and ours, which natively encode absolute timestamps, we additionally append the frame timestamps to the text input. To ensure a fair comparison, all VLMs are instructed to directly output final choice without intermediate reasoning. For evaluation on Video-MME \cite{fu2025video}, video subtitles are excluded from input.

\subsection{QA Generation Details}
\noindent \textbf{Object Denotation.} In both DSR-Train and DSR-Bench, each object is referenced using a combination of its category and bounding box coordinates either at a specific timestamp or at the beginning of the queried sub-interval. This differs from some prior benchmarks that refer to objects solely through appearance descriptions, which can be ambiguous in our setting where many objects share similar visual characteristics. When an object is used to determine the absolute viewpoint, it will be denoted as ``\{class\} with bounding box coordinates ($\text{x}_1,\text{y}_1,\text{x}_2,\text{y}_2$) at \{$\text{time}_\text{v}$\}'', where ``\{class\}'' is the object category and ``($\text{x}_1,\text{y}_1,\text{x}_2,\text{y}_2$)'' correspond to the top-left and bottom-right coordinates of the bounding box in the frame sampled at the timestamp ``\{$\text{time}_\text{v}$\}''. Otherwise, the object is referred to as ``\{class\} with initial bounding box coordinates ($\text{x}_1,\text{y}_1,\text{x}_2,\text{y}_2$), indicating that the coordinates correspond to the first frame of the queried sub-interval and can change thereafter, or as ``\{class\} with final bounding box coordinates ($\text{x}_1,\text{y}_1,\text{x}_2,\text{y}_2$)'' to indicate that the coordinates are taken from the last sampled video frame and the task is to predict object's moving direction.

\begin{figure*}[t]
\centering
\setlength{\belowcaptionskip}{-5pt}
\includegraphics[width=0.95\linewidth]{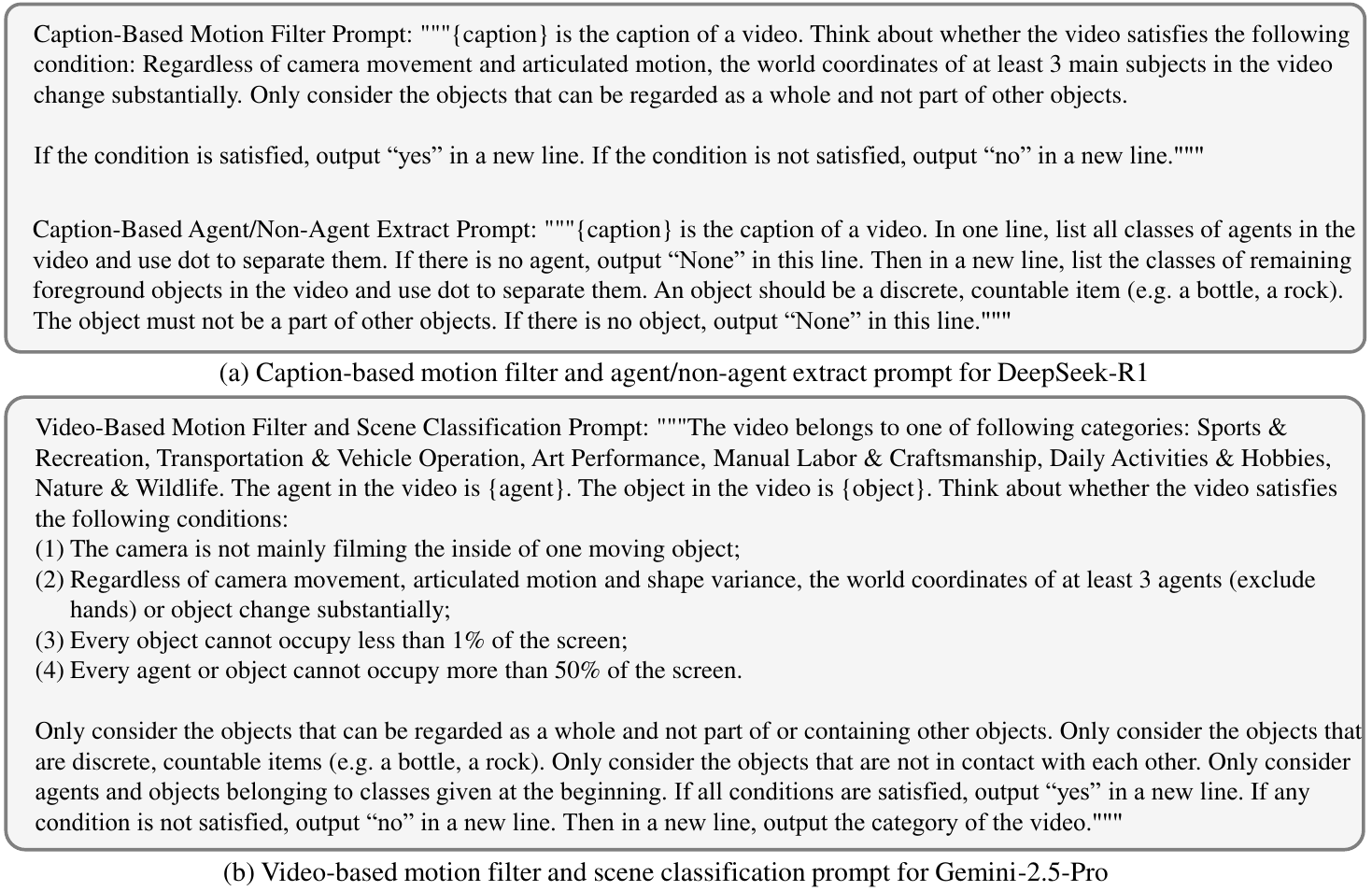}
\caption{Detailed prompts leveraged for DeepSeek-R1 and Gemini-2.5-Pro for data process.}
\label{img:s2}
\end{figure*}

\noindent \textbf{Prompt Design.} As described in Section \ref{sec:3.1} and \ref{sec:3.2} of the main manuscript, our automated QA generation pipeline uses DeepSeek-R1 \cite{guo2025deepseek} to filter out Koala-36M \cite{wang2025koala} videos exhibiting negligible object motion and to identify agent and non-agent object categories based on the video captions. To construct DSR-Bench, we further employ Gemini-2.5-Pro \cite{comanici2025gemini} to filter videos with visual content and to assign scene classes with higher accuracy. Figure \ref{img:s2} presents our prompts for DeepSeek-R1 and Gemini-2.5-Pro, where “\{caption\}” denotes the video caption, and “\{agent\}” and “\{object\}” are agent and non-agent classes in the video.

\begin{figure*}[t]
\centering
\setlength{\belowcaptionskip}{-10pt}
\includegraphics[width=0.95\linewidth]{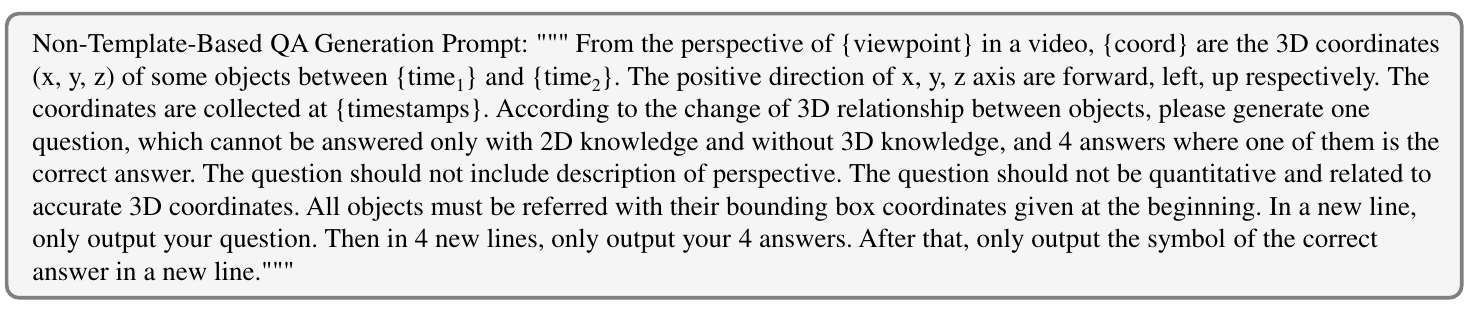}
\caption{Detailed prompts leveraged for DeepSeek-R1 to generate non-template-based QAs.}
\label{img:s3}
\end{figure*}

In Section \ref{sec:3.3} of the main manuscript, to broaden linguistic variety and reasoning patterns, we also propose to prompt DeepSeek-R1 for non-template-based QA generation. In Figure \ref{img:s3}, we show the detailed prompt for DeepSeek-R1 for QA generation, where “\{viewpoint\}” is the observer to decide the viewpoint for observing the objects, “\{coord\}” is the 3D coordinate trajectories of all objects, “$\{\text{time}_1\}$” and “$\{\text{time}_2\}$” are the start and end timestamp of the sub-interval, “\{timestamps\}” is the list of timestamps at which the object 3D coordinates are collected.

\noindent \textbf{Question Generation Template.} Part of our QAs are generated based on pre-defined templates, as stated in Section \ref{sec:3.3} of the main manuscript. We list these templates for each type of question in Table \ref{tab:s1}. “\{$\text{time}_\text{s}$\}”, “\{$\text{time}_\text{e}$\}” are the timestamps of the start, end of the queries sub-interval and “\{$\text{time}_\text{v}$\}” is the timestamp to determine the absolute viewpoint. “\{$\text{obj}_\text{1}$\}”, “\{$\text{obj}_\text{2}$\}” are
 the target objects and “\{$\text{obj}_\text{v}$\}” is the observer to decide the viewpoint, where their denotations are in the format described in the previous \textbf{Object Denotation} section.
 
\noindent \textbf{Answer Generation Rule.} In Section \ref{sec:3.3} of the main manuscript, we describe our approach to generate answers by comparing the queried attribute every 2 adjacent frames to derive a sequence of basic choice defined in Table \ref{tab:2} of the main manuscript and then merging consecutive identical states. Table \ref{tab:s2} provides the detailed derivation rules for each basic choice. After determining the correct answer, we compute the number of basic choices
$N$ it contains and generate alternative choices with random lengths within the range [$\text{max}(1,N-3), N+3$]. To maintain a balanced distribution, the correct answer label (A, B, C, or D) is randomly assigned. We note that frames in the final second are hidden from VLMs, as they are used to determine the object’s movement direction for direction prediction questions.

\onecolumn
\begin{table*}[h!]
\caption{Templates for different question types.}
\centering
\renewcommand{\arraystretch}{1.3}
\resizebox{0.8\width}{!}{
\begin{tabular}{cc}
\shline
Type                                                   & Template                                                                                                         \\ \shline
Rel Dis                                                & Between \{$\text{time}_\text{s}$\} and \{$\text{time}_\text{e}$\}, following the perspective of \{$\text{obj}_\text{v}$\}, how does the distance between \{$\text{obj}_\text{1}$\} and \{$\text{obj}_\text{2}$\} change?    \\ 
Rel Dir                                                & Between \{$\text{time}_\text{s}$\} and \{$\text{time}_\text{e}$\}, following the perspective of \{$\text{obj}_\text{v}$\}, how does the direction of \{$\text{obj}_\text{1}$\} to \{$\text{obj}_\text{2}$\} change?         \\ 
Rel Ori                                                & Between \{$\text{time}_\text{s}$\} and \{$\text{time}_\text{e}$\}, following the perspective of \{$\text{obj}_\text{v}$\}, how does the orientation of \{$\text{obj}_\text{1}$\} change?               \\ 
Rel Spd                                                & Between \{$\text{time}_\text{s}$\} and \{$\text{time}_\text{e}$\}, following the perspective of \{$\text{obj}_\text{v}$\}, how does the speed of \{$\text{obj}_\text{1}$\} change?                     \\ 
Rel Spd Comp & Between \{$\text{time}_\text{s}$\} and \{$\text{time}_\text{e}$\}, following the perspective of \{$\text{obj}_\text{v}$\}, compare the speed between \{$\text{obj}_\text{1}$\} and \{$\text{obj}_\text{2}$\}.               \\ 
Rel Dir Pred & Following the perspective of \{$\text{obj}_\text{v}$\}, predict the moving direction of \{$\text{obj}_\text{1}$\}.                                         \\ 
Abs Dis                                                & Between \{$\text{time}_\text{s}$\} and \{$\text{time}_\text{e}$\}, from the perspective of \{$\text{obj}_\text{v}$\} at \{$\text{time}_\text{v}$\}, how does the distance between \{$\text{obj}_\text{1}$\} and \{$\text{obj}_\text{2}$\} change? \\ 
Abs Dir                                                & Between \{$\text{time}_\text{s}$\} and \{$\text{time}_\text{e}$\}, from the perspective of \{$\text{obj}_\text{v}$\} at \{$\text{time}_\text{v}$\}, how does the direction of \{$\text{obj}_\text{1}$\} to \{$\text{obj}_\text{2}$\} change?      \\ 
Abs Ori                                                & Between \{$\text{time}_\text{s}$\} and \{$\text{time}_\text{e}$\}, from the perspective of \{$\text{obj}_\text{v}$\} at \{$\text{time}_\text{v}$\}, how does the orientation of \{$\text{obj}_\text{1}$\} change?            \\ 
Abs Spd                                                & Between \{$\text{time}_\text{s}$\} and \{$\text{time}_\text{e}$\}, from the perspective of \{$\text{obj}_\text{v}$\} at \{$\text{time}_\text{v}$\}, how does the speed of \{$\text{obj}_\text{1}$\} change?                  \\ 
Abs Spd Comp & Between \{$\text{time}_\text{s}$\} and \{$\text{time}_\text{e}$\}, from the perspective of \{$\text{obj}_\text{v}$\} at \{$\text{time}_\text{v}$\}, compare the speed between \{$\text{obj}_\text{1}$\} and \{$\text{obj}_\text{2}$\}.            \\ 
Abs Dir Pred & From the perspective of \{$\text{obj}_\text{v}$\} at \{$\text{time}_\text{v}$\}, predict the moving direction of \{$\text{obj}_\text{1}$\}.                                      \\ \shline
\end{tabular}
}
\label{tab:s1}
\end{table*}
\begin{table*}[h!]\small
\caption{Derivation rules of different basic choices to make up answers.}
\centering
\renewcommand{\arraystretch}{1.3}
\begin{tabularx}{0.98\textwidth}{ccX}
\shline
Type                                                                                                        & Basic Choice                                 & ~~~~~~~~~~~~~~~~~~~~~~~~~~~~~~~~Derivation Rule                                                                                                                                                                                                                                   \\ \shline
\multirow{5}{*}{\begin{tabular}[c]{@{}c@{}}\\\\\\\\\\Distance \& Speed \end{tabular}}                                                                          & (1) Keep nearly constant then become larger  & At each timestamp within a period, except final one, the distance/speed remains within 0.8$\times$ to 1.2$\times$ that of the first timestamp. At the final timestamp, the distance/speed exceeds 1.2$\times$  that of the first timestamp.     \\
                                                                                                            & (2) Keep nearly constant then become smaller & At each timestamp within a period, except final one, the distance/speed remains within 0.8$\times$ to 1.2$\times$ that of the first timestamp. At the final timestamp, the distance/speed is below 0.8$\times$  that of the first timestamp. \\
                                                                                                            & (3) Keep nearly constant                     & At each timestamp within a period, except final one, the distance/speed remains within 0.8$\times$ to 1.2$\times$ that of the first timestamp.                                                                                          \\
                                                                                                            & (4) Become larger                            & At each timestamp within a period, the distance/speed exceeds 1.2$\times$ that of the former one.                                                                                                                                         \\
                                                                                                            & (5) Become smaller                           & At each timestamp within a period, the distance/speed falls below 0.8$\times$ that of the former one.                                                                                                                                     \\ \hline
\multirow{3}{*}{\begin{tabular}[c]{@{}c@{}}\\\\\\Direction \& Orientation\\ \& Direction Prediction\end{tabular}} & (1) Front/Behind                             & The angle between the unit vector of direction/orientation and the forward unit vector from viewpoint is smaller/larger than 70/110 degrees.                                                                                                                       \\
                                                                                                            & (2) Left/Right                               & The angle between the unit vector of direction/orientation and the left unit vector from viewpoint is smaller/larger than 70/110 degrees.                                                                                                                          \\
                                                                                                            & (3) Above/Below                              & The angle between the unit vector of direction/orientation and the upward unit vector from viewpoint is smaller/larger than 70/110 degrees.                                                                                                                        \\ \hline
\multirow{3}{*}{\begin{tabular}[c]{@{}c@{}}\\\\Speed Comparison\end{tabular}}                                                                           & (1) Nearly the same                          & The speed of the former object is within 0.83$\times$ to 1.20$\times$ that of the latter object at one timestamp.                                                                                                                                                  \\
                                                                                                            & (2) The former is faster                     & The speed of the former object exceeds 1.20$\times$ that of the latter object at one timestamp.                                                                                                                                                            \\
                                                                                                            & (3) The latter is faster                     & The speed of the former object falls below 0.83$\times$ that of the latter object at one timestamp.                                                                                                                                                        \\ \shline
\end{tabularx}
\label{tab:s2}
\end{table*}
\twocolumn

\section{Complementary Experiments}
\label{sec:s2}
In this section, we present additional experiments that were not included in the main manuscript. In Section \ref{sec:s2.3}, we provide further analysis of the results in Table \ref{tab:4} of main manuscript. Section \ref{sec:s2.1} compare the performance of models trained on data with different question type distributions. In Section \ref{sec:s2.2}, we replace the original base model, \ie, Qwen2.5-VL-7B, with Qwen3-VL-8B and report performance across different benchmarks to demonstrate the effectiveness of GSM and DSR-Train on different models. Section \ref{sec:8.4} incorporates DSR-Train with question–answer pairs for spatial reasoning in static scenes to train Qwen2.5-VL-7B with unified static and dynamic spatial reasoning capabilities. Finally, Section \ref{sec:8.5} extends the DSR-Train fine-tuned model fine-tuned to MineDojo \cite{fan2022minedojo}, demonstrating its broader applicability to downstream agent tasks requiring dynamic spatial reasoning.

\subsection{Additional Result Analysis}
\label{sec:s2.3}

In addition to the conclusions in Section 5.2, Table \ref{tab:4} of the main manuscript further reveals that spatial reasoning models, \ie, VLM-3R \cite{fan2025vlm} and VG-LLM \cite{zheng2025learning}, achieve even better performance than proprietary models, despite being trained only for static spatial reasoning. Therefore, 3D relevant data is necessary to improve spatial reasoning capability. Moreover, compared with its base model LLaVA-Video-7B, the marginal performance gain of VideoRefer \cite{yuan2025videorefer}, which is explicitly trained to understand location information for object reference, suggests that the unsatisfactory performance on DSR-Bench is not attributable to our use of bounding-box–based object references. Instead, it highlights the insufficient capability of current models to perform dynamic spatial reasoning.

\subsection{Effect of Training Question Distribution}
\label{sec:s2.1}
For all experiments in the main manuscript, our model is trained on QAs with the type distribution shown in Figure \ref{img:s1}. In this section, we fix the total number of training QAs to be 20K and vary the proportion of template-based and non-template-based QAs under the following configurations: (1) 20K template-based QAs; (2) 16K template-based QAs and 4K non-template-based QAs (the setting adopted in the main manuscript experiments); (3) 10K template-based QAs and 10K non-template-based QAs; (4) 20K non-template-based QAs. Table \ref{tab:s3} reports the performance on the template-based subset, non-template-based subset, the full DSR-Bench and Video-MME under each configuration. The results indicate that both QA types are essential for achieving strong overall performance, and that template-based QAs should constitute the majority.

\begin{table}[h]
\caption{Comparison between different training data settings.}
\centering
\renewcommand{\arraystretch}{1.3}
\resizebox{0.81\width}{!}{
\begin{tabular}{c|cccc}
\shline
\multirow{2}{*}{\begin{tabular}[c]{@{}c@{}}\\Settings\end{tabular}} & \multicolumn{4}{c}{Benchmarks}                                                                                                                                  \\ \cline{2-5} 
                          & \begin{tabular}[c]{@{}c@{}}DSR-Bench\\ (Temp)\end{tabular} & \begin{tabular}[c]{@{}c@{}}DSR-Bench\\ (Non-Temp)\end{tabular} & DSR-Bench & Video-MME \\ \shline
Baseline                  & 22.1                                                             & 30.1                                                                 & 23.5      & 60.2      \\
1                         & 61.5                                                             & 32.3                                                                 & 56.2      &      59.6     \\
2                         & 60.4                                                             & 43.7                                                                 & 57.4      & 59.9      \\
3                         & 58.1                                                             & 46.7                                                                 & 56.1      &      60.1     \\
4                         &  35.2                                                                &      49.1                                                                &     37.6      &     60.5      \\ \shline
\end{tabular}
}
\label{tab:s3}
\end{table}

\subsection{Ablation on Different Base Model}
\label{sec:s2.2}
In the main manuscript experiments, we adopt Qwen2.5-VL-7B as the base model, integrate our proposed GSM into it and train it on DSR-Train. To demonstrate the effectiveness of GSM and DSR-Train independent of the underlying base model, we further replace Qwen2.5-VL-7B with Qwen3-VL-8B-Instruct, incorporate GSM, and train it on the same 20K samples from DSR-Train as used in Section \ref{sec:5.3} of the main manuscript. We also compare its performance with different training methods when adopting the same compared training paradigm in Section \ref{sec:5.3} of main manuscript, \ie, SFT and Addition. The results on DSR-Bench, STI-Bench \cite{li2025sti}, VLM4D \cite{zhou2025vlm4d} and Video-MME \cite{fu2025video} listed in Table \ref{tab:s4} show that our DSR-Train improves the dynamic spatial reasoning performance of Qwen3-VL-8B-Instruct as well. Compared with SFT, GSM further improves performance by integrating explicit geometric priors. In contrast to Addition, GSM maintains the general video-understanding ability of Qwen3-VL-8B-Instruct, preventing degradation of broader visual reasoning skills.

\begin{table}[t]
\caption{Comparison between GSM and other methods with Qwen3-VL-8B-Instruct as the base model.}
\centering
\renewcommand{\arraystretch}{1.3}
\resizebox{0.78\width}{!}{
\begin{tabular}{c|ccccc}
\shline
\multirow{2}{*}{Methods}                              & \multicolumn{5}{c}{Benchmarks}                   \\ \cline{2-6} 
                                                      & DSR-Bench & VLM4D & STI-Bench & Video-MME & Avg. \\ \shline
Baseline                                              &      28.7     &    46.3   &     36.3      &     64.9      &   44.0   \\
SFT                                                   &     56.8      &    47.8   &     37.3      &     64.7      &   51.6   \\
Addition                                                   &    59.0       &    49.5   &     37.8      &     53.6      &   49.9   \\
GSM                                                  &     58.6      &    49.2   &     37.9      &      64.4     &   52.5   \\ \shline
\end{tabular}
}
\label{tab:s4}
\end{table}

\begin{figure*}[t]
\centering
\includegraphics[width=\linewidth]{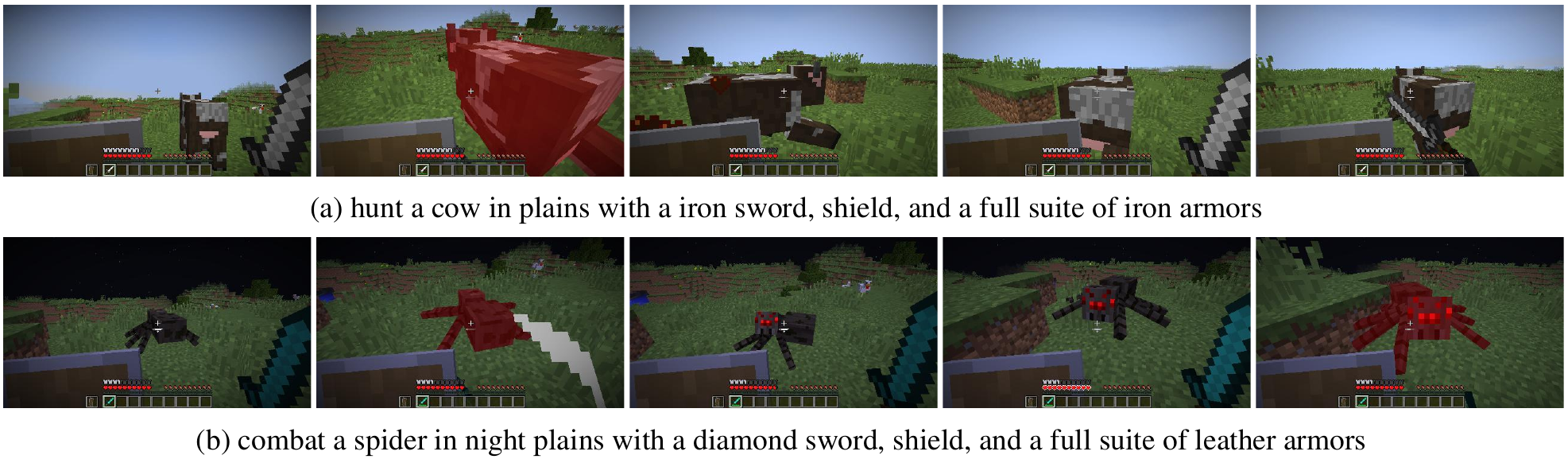}
\caption{Examples of agent results on MineDojo.}
\label{img:s7}
\end{figure*} 

\subsection{Performance with Data Mixture}
\label{sec:8.4}
In all previous experimental settings, the base model Qwen2.5-VL-7B is fine-tuned exclusively on our DSR-Train to perform spatial reasoning in dynamic scenes. In this section, we further mix QAs for static spatial reasoning with DSR-Train to fine-tune Qwen2.5-VL-7B, improving both static and dynamic spatial reasoning ability. Specifically, we construct a total of 800K static spatial reasoning QAs using data from VLM-3R \cite{fan2025vlm} and Cambrian-S \cite{yang2025cambrian}, and adopt VSI-Bench \cite{yang2025thinking} as the benchmark for evaluating static spatial reasoning. The performance of model trained only on static spatial reasoning or mixed QAs on DSR-Bench, VSI-Bench and Video-MME are listed in Table \ref{tab:s5}. From the results, when the model is trained only with static spatial reasoning QAs, it is still unable to perform dynamic spatial reasoning. With both static and dynamic spatial reasoning data, the trained model obtains the best performance simultaneously on DSR-Bench and VSI-Bench while general understanding capability is still preserved.

\begin{table}[t]
\caption{Performance analysis of models trained with different spatial reasoning scenes.}
\centering
\renewcommand{\arraystretch}{1.3}
\resizebox{0.81\width}{!}{
\begin{tabular}{c|ccc}
\shline
\multirow{2}{*}{Training Scene} & \multicolumn{3}{c}{Benchmarks}                                                                                                                                  \\ \cline{2-4} 
                          & DSR-Bench & VSI-Bench & Video-MME \\ \shline
Baseline                  & 23.5                                                                 & 33.4      & 60.2      \\
Static                         & 24.1          & 55.3      &      58.9     \\
Static+Dynamic                         &  60.2        &     56.1      &     59.2      \\ \shline
\end{tabular}
}
\label{tab:s5}
\end{table}

\subsection{Extension to Agent Tasks}
\label{sec:8.5}
In this section, we also explore the downstream performance on agent tasks~\cite{lin2025embrace,fan2022minedojo}, showing the spatial capability on real-time working agent. We further extend the Qwen2.5-VL-7B fine-tuned with DSR-Train to MineDojo \cite{fan2022minedojo} Benchmark, which simulates a Minecraft game agent to solve different tasks, to show the application of dynamic spatial reasoning models in downstream agent tasks. Since the training dataset of MineDojo only provides videos and task instructions without intermediate action labels, we instead leverage the data from MineRL \cite{guss2019minerl} for training by mapping its action space to that of MineDojo. For evaluation, we select part of tasks from MineDojo that are related to the following two sets of objects: (1) Animals including cow, sheep, pig and chicken, whose related tasks are targeted at combating or harvesting them; (2) Hostiles including spider, zombie, pigman and enderman, whose related tasks aim to combat them. Since these tasks require interacting with dynamic objects, they command dynamic spatial reasoning ability of agents. For compared models, we train original Qwen2.5-VL-7B and that fine-tuned with static spatial reasoning data as stated in Section \ref{sec:8.4} on the same MineRL data. Each task is tested on 3 random seeds and with 200 episodes. The average and variance of success rate are listed in Table \ref{tab:s6}, showing the advantage of our model in downstream agent tasks. Some task results of the agent adopted from model trained with DSR-Train are illustrated in Figure \ref{img:s7}.
\begin{table}[t]
\caption{Success rate comparison on MineDojo between agents adopted from VLM trained with different spatial reasoning scenes.}
\centering
\renewcommand{\arraystretch}{1.3}
\resizebox{0.81\width}{!}{
\begin{tabular}{c|cc}
\shline
\multirow{2}{*}{Training Scene} & \multicolumn{2}{c}{Tasks}                                                                                                                                  \\ \cline{2-3} 
                          & Animals & Hostiles \\ \shline
Baseline                  & 15.6$\pm$3.4      & 10.2$\pm$2.4      \\
Static                         & 16.3$\pm$2.3      & 12.4$\pm$3.1     \\
Dynamic                         &  26.5$\pm$1.7     & 22.3$\pm$2.3      \\ \shline
\end{tabular}
}
\label{tab:s6}
\end{table}

\section{Visualization}
\label{sec:s3}
\begin{figure*}[t]
\centering
\includegraphics[width=0.92\linewidth]{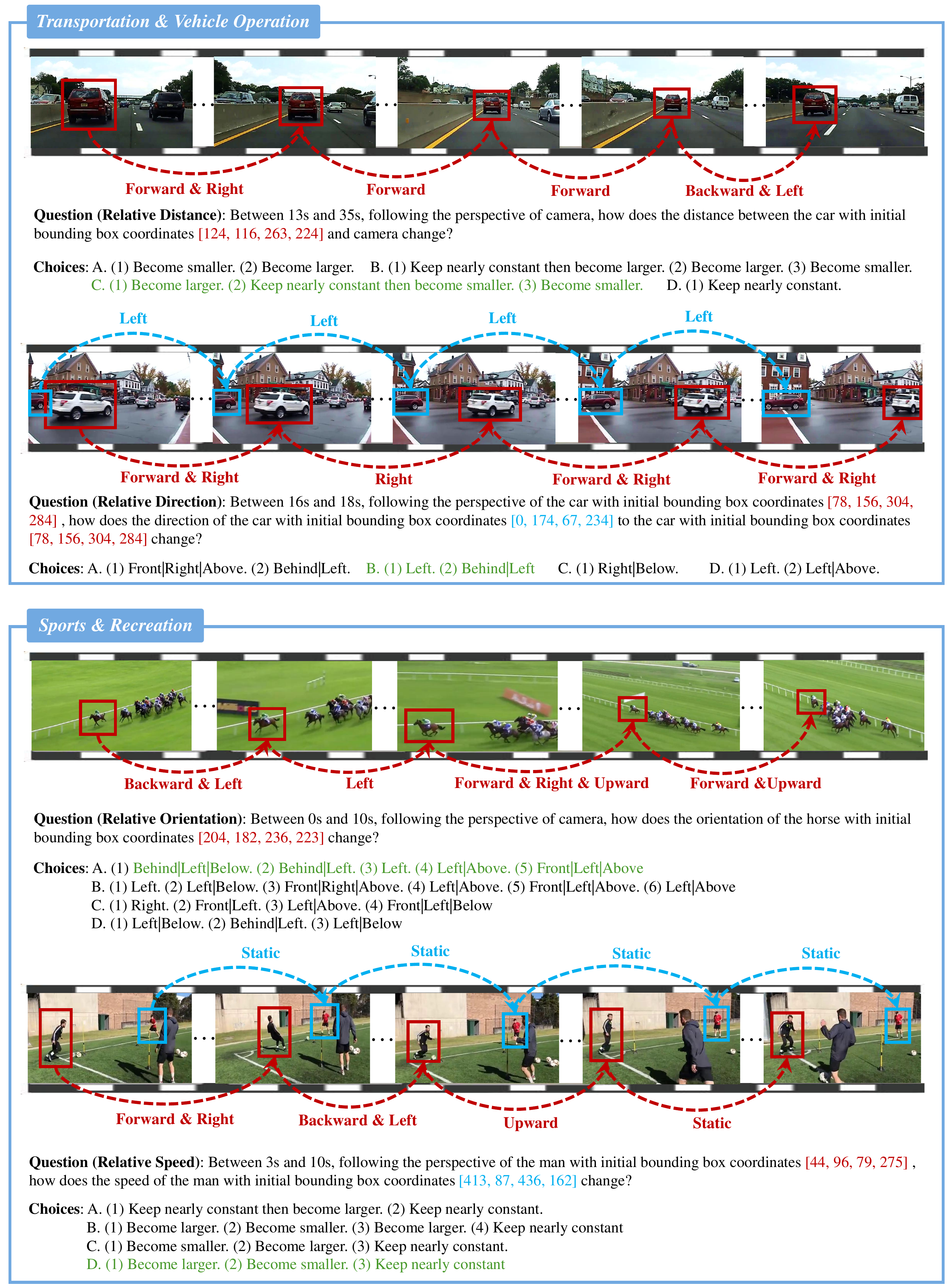}
\caption{Examples of QA and video in DSR-Train and DSR-Bench.}
\label{img:s4}
\end{figure*}

\begin{figure*}[t]
\centering
\includegraphics[width=0.95\linewidth]{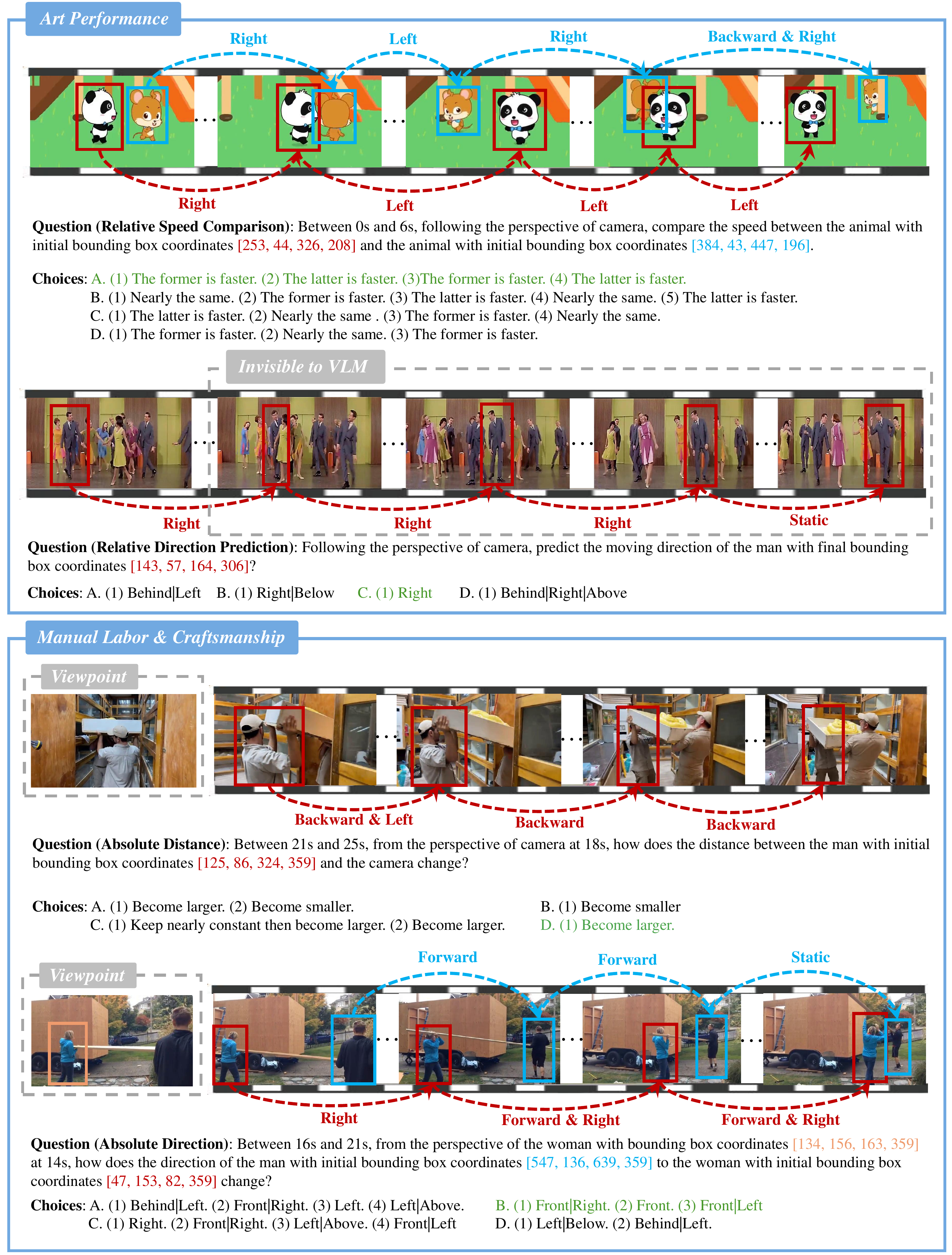}
\caption{Examples of QA and video in DSR-Train and DSR-Bench.}
\label{img:s5}
\end{figure*}

\begin{figure*}[t]
\centering
\setlength{\abovecaptionskip}{3pt}
\includegraphics[width=0.87\linewidth]{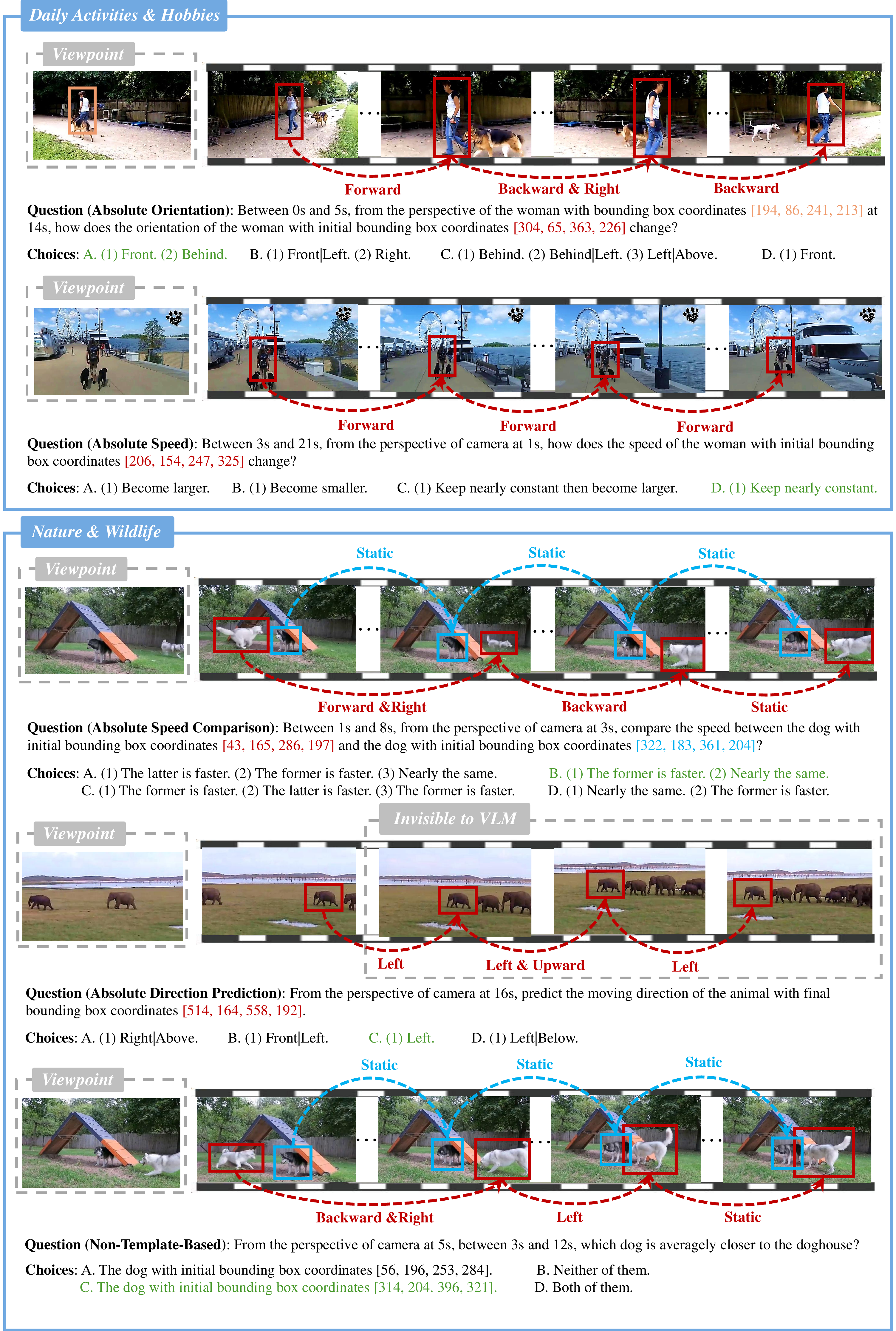}
\caption{Examples of QA and video in DSR-Train and DSR-Bench.}
\label{img:s6}
\end{figure*}

In this section, we illustrate the video and QA examples of different categories in our DSR-Train and DSR-Bench in Figure \ref{img:s4}, \ref{img:s5} and \ref{img:s6}, which shows our in-the-wild video source and comprehensive evaluation aspects including multi-object interaction, viewpoint transformation and fine-grained temporal reasoning. Note that the arrows in the figures indicate the movement from the viewpoint of camera, which can be different when the viewpoint changes.

\end{document}